\documentclass[10pt,journal,compsoc]{IEEEtran}

\usepackage{graphicx}
\usepackage{soul}
\usepackage{color}
\usepackage{algorithm}
\usepackage{algorithmic}
\usepackage{enumitem}
\usepackage[caption=false]{subfig}
\usepackage{amsfonts}

\ifCLASSOPTIONcompsoc
  \usepackage[nocompress]{cite}
\else
  \usepackage{cite}
\fi
\ifCLASSINFOpdf
\else
\fi

\begin{document}

%\title{In-Flight Energy Recharging of Drone Swarms for Delivery Services}
\title{In-Flight Energy-Driven Composition of Drone Swarm Services}


% \author{\IEEEauthorblockN{Balsam Alkouz\IEEEauthorrefmark{1},
% Amani Abusafia\IEEEauthorrefmark{1},
% Abdallah Lakhdari\IEEEauthorrefmark{1}, and
% Athman Bouguettaya\IEEEauthorrefmark{1}
% }\\
% \IEEEauthorblockA{\IEEEauthorrefmark{1}School of Computer Science \\The University of Sydney\\ Sydney, NSW, Australia\\
% Emails: \{balsam.alkouz,  amani.abusafia, abdallah.lakhdari, athman.bouguettaya\}@sydney.edu.au}
% }

\author{Balsam~Alkouz,
        Amani~Abusafia,
        Abdallah~Lakhdari,
        and~Athman~Bouguettaya,~\IEEEmembership{Fellow,~IEEE}% <-this % stops a space
\IEEEcompsocitemizethanks{\IEEEcompsocthanksitem B. Alkouz, A. Abusafia, A. Lakhdari, A. Bouguettaya are with the School of Computer Science,
University of Sydney, Australia.
E-mail: \{balsam.alkouz, amani.abusafia, abdallah.lakhdari, athman.bouguettaya\}@sydney.edu.au
%\protect\\
%E-mail: see http://www.michaelshell.org/contact.html
}% <-this % stops an unwanted space
% \thanks{Manuscript received April 19, 2005; revised August 26, 2015.}
}

% \markboth{Journal of \LaTeX\ Class Files,~Vol.~14, No.~8, August~2015}%
% {Shell \MakeLowercase{\textit{et al.}}: Bare Demo of IEEEtran.cls for Computer Society Journals}

\IEEEtitleabstractindextext{%
\begin{abstract}
We propose a novel framework for swarm-based drone delivery services with in-flight energy recharging. The framework aims to enhance the delivery time of multiple packages by reducing the number of stops and recharging times at intermediate stations. The proposed framework considers various \textit{intrinsic and extrinsic} delivery constraints. We propose to use \textit{support drones} whose sole purpose is to recharge other drones in the swarm during their flight. 
%We propose a framework where the optimal number of support drones in a swarm is determined %by a \textit{redundancy-based} method to minimize the probability of delivery failures. 
In this respect, we compute the optimal set of optimal support drones to minimize the probability of delivery services and recharging time at the next stations. We also use two settings to position the support drones in a flight formation for comparative purposes. Two novel \textit{energy sharing} methods are proposed, namely, Priority-based and Fairness-based methods. A re-ordering method of the delivery drones is presented to facilitate the in-flight energy composition process. An enhanced A* algorithm is implemented to compose the optimal services in terms of delivery time. Experimental results prove the efficiency of our proposed approach.
\end{abstract}

\begin{IEEEkeywords}
IoT services, Drones swarm, In-flight energy recharging, Service composition.
\end{IEEEkeywords}}

\maketitle

\IEEEdisplaynontitleabstractindextext
\IEEEpeerreviewmaketitle


% \fi
\IEEEpeerreviewmaketitle

\section{Introduction}

Combining unmanned aerial vehicle technologies with the service paradigm has given rise to the concept of \textit{Drone services} \cite{hamdi2021drone}. It is defined as the use of drones to provide services \cite{hamdi2021drone}.  A drone has the ability to sense the surrounding environment and carry payloads. DaaS are now routinely used in such domains as  traffic monitoring \cite{sharma2018lorawan}, agriculture \cite{skobelev2018designing}, and delivery \cite{hamdi2021drone}. Drone \textit{delivery services} are the new technology that promises impactful and innovative solutions to the ever expanding online shopping  \cite{alkouz2022density}. Drones provide a more cost-effective and environmentally friendly alternative when compared to \textcolor{black}{terrestrial delivery services like trucks \cite{alkouz2021service}. Examples of terrestrial services include Fedex and UPS.} In addition, drones have been an enabling technology to mitigate the effect of pandemics such as COVID-19 by creating more resilient supply chains and socially distanced delivery services \cite{euchi2020drones}. In that regard, the use of drones allows for the autonomous  delivery of goods in a \textit{quicker}, \textit{cheaper}, and \textit{safer} i.e., contactless manner. Major retailers are already integrating drones within their delivery options after governments around the world started relaxing regulations in response to the COVID-19 pandemic. For instance, Google’s Wing has expanded its drone delivery services by doubling its deployment rates in Australia\footnote{https://www.afr.com/technology/google-spreads-wings-as-drone-deliveries-up-500pc-20200929-p560bw}. 

While single drone delivery provides a multitude of opportunities, there are instances where a \emph{swarm} of drones may be required to fulfill the requirements of a delivery request. These include instances where a package is too heavy to be delivered by any one single drone or when multiple packages need to be delivered together. Federal aviation regulations dictate that payloads per drone not exceed 2.5kg\footnote{Federal Aviation Administration (FAA) in USA- https://www.faa.gov/uas/advanced\_operations/package\_delivery\_drone}\textsuperscript{,}\footnote{\textcolor{black}{Drone delivery company confident of taking flight in Queensland, Australia- https://www.brisbanetimes.com.au/\\national/queensland/drone-delivery-company-confident-of-taking-flight-in-queensland-20190731-p52ckx.html}}. Hence, the swarm-based drone services are an effective alternative and solution to address these limitations \cite{alkouz2020swarm}. The \textit{focus of this research is on the use of drone swarms for the delivery of packages as a service}. More formally, a swarm is defined as a set of drones that move together within close proximity and arriving at any stop within a time window, thus acting as a single entity to achieve a certain goal. Swarms of drones have been used in various applications including military \cite{wilkerson2016aerial}, airborne communication networks \cite{shi2018drone}, sky shows \cite{waibel2017drone}, and delivery \cite{alkouz2021provider}. 

%In delivery, a scenario could happen when a requested package exceeds a single drone payload capacity due to strict federal aviation regulations that approve only small drones for delivery in the city (payload $<$ 2.5kg)\footnote{https://www.faa.gov/uas/advanced\_operations/package\_delivery\_drone}. In this case, a divided package could be carried by multiple drones in a swarm. A swarm is also effective in distributing the energy burden amongst the drones by distributing the payload, which allows the swarm to travel to farther destinations. However, these advantages come with the cost of the multiple constraints a swarm would face during the delivery. These constraints may include the delimited arrival time of all the drones in the swarm to the destination. In addition, drones in a swarm would usually carry different payloads, which causes different rates of energy consumption within a swarm.

The \textit{limited battery capacity} of drones presents a major challenge in swarm-based drone delivery services. \textcolor{black}{Several solutions have been proposed to address this limitation. Intuitively, battery replacement would be the obvious solution. %However, this would likely require human intervention. If we are aiming at a fully autonomous drone service, this would be undesirable. 
However, most automatic battery replacement approaches exhibit inadequate accuracy in outdoor environments \cite{fujii2013endless}. Carrying additional batteries may be another solution. However, given the drones' small size, this would severely limit their useful load. Solar energy is another option.  However, its dependence on the weather conditions and slow charging process would make it hardly practical \cite{galkin2019uavs}. Moreover, the size of the solar panels would add to the payload \cite{galkin2019uavs}. Another suggested solution in the literature is the use of laser beams fitted at rooftops. This would not be applicable in a delivery scenario as the range of the beam will vary when the drones move. Furthermore, the line-of-sight would be blocked by the urban environment \cite{galkin2019uavs}.} A more realistic option is the use of \textit{recharging pads} at delivery/recharging stations \cite{alkouz2020swarm}. A major challenge is the limited number of recharging stations, thus limiting the number of parallel recharging of drones. This may result in extra waiting times until all drones in a swarm are recharged as they act as an atomic unit for delivery purposes. Consequently, delivery times would likely be negatively impacted. \emph{We propose the use of in-flight recharging drones to speed up swarm-based drone package delivery services}. We introduce the concept of \emph{support drones} that fly with a swarm to in-flight charge other drones.

In-flight charging refers to the technology that enables recharging drones' batteries on the fly. For example, GET Air\footnote{ Global Energy Transmission: http://getcorp.com/technology-overview/} solution provides an in-flight wireless power charger which can give a maximum output power of 12kW. The drone only needs to hover over a recharging power station for six minutes. A new approach allows drones to deliver power wirelessly while in-flight \cite{benbuk2020leveraging}. The drone, in this case, transmits signals that can be harvested and rectified to DC voltage. We focus on \textit{energy sharing services}, through the use of in-flight recharging, as an overlay to expedite swarm-based drone service delivery. Energy sharing services allow drone swarms to travel farther destinations reducing the number of stops needed to recharge. 
\emph{Energy sharing} refers to the general process of sharing energy among IoT devices \cite{lakhdari2020composing}. An IoT device (in our case, a drone) that shares energy is called an \emph{energy provider}. In contrast, an IoT device that requires energy is called an \emph{energy consumer}. In a swarm-based drone delivery services context, we view a drone that transmits energy as the provider. The consumer is the drone that  received the energy.

Swarm-based drone delivery services lend themselves quite naturally to being modeled using the service paradigm because they map to the key ingredients of the services' concept, i.e., functional and non-functional attributes \cite{alkouz2020swarm}. The \textit{function} of a swarm-based drone service is the delivery of packages from a source to a destination by a swarm. The \textit{non-functional} properties include energy consumption, delivery time, etc. Likewise, we leverage the service paradigm to model energy sharing services, termed as \emph{Energy-as-a-Service (EaaS)} \cite{lakhdari2020composing}\cite{lakhdari2018crowdsourcing}. The function of EaaS is to share energy between a provider and a consumer. The non-functional properties include the amount of energy, sharing duration, etc. We propose \textit{a nested two-level service composition} framework that consists of composing swarm-based drone services while achieving an \textcolor{black}{Energy-as-a-Service} composition (EaaS).  In this respect, the first level composition of energy services acts as a non-functional (QoS) for the second level of composition of swarm-based services.  Therefore, finding an optimal plan for swarm-based drone services for delivering goods from a source to a destination is a two-step service composition: 1. Finding the best energy service composition between each two directly connected node, and 2. Finding the best path using other criteria (e.g., time, cost, etc) between source and destination. The main contributions of this work are as follows:
\begin{itemize}
    \item A novel in-flight energy sharing model in swarm-based drone delivery services.
    \item{A Constraint-aware swarm-based drone services composition using  a  modified  A* heuristic algorithm  to  compose the optimal delivery services.}
    \item A two-level composition framework where the first level is the composition of energy services and second level is the composition of swarm-based services.
    
\end{itemize}

% the structure of the paper

\section{Motivating Scenario}

\begin{figure}[!t]
\centering
  \centering
  \includegraphics[width=\linewidth]{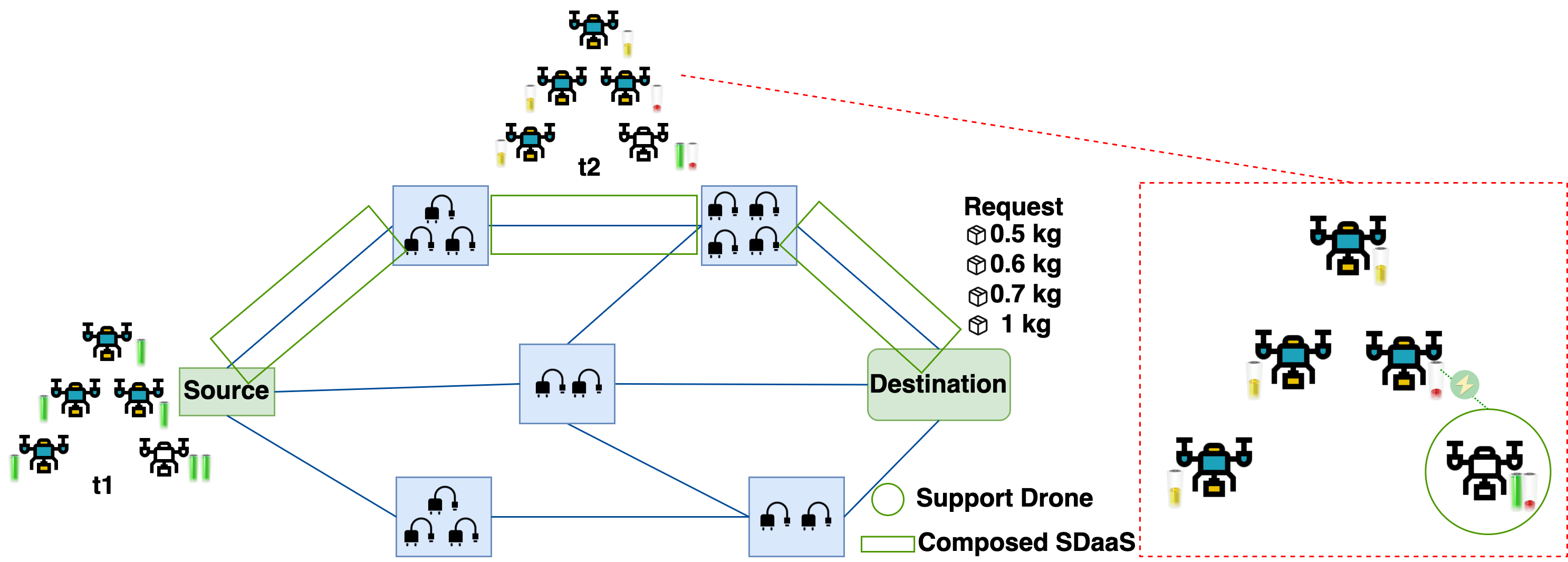}
  \caption{Swarm-Based Delivery Services with In-Flight Energy Sharing}
  \label{fig:motivating}
\end{figure}

\begin{figure}[!t]
\centering
  \centering
  \includegraphics[width=\linewidth]{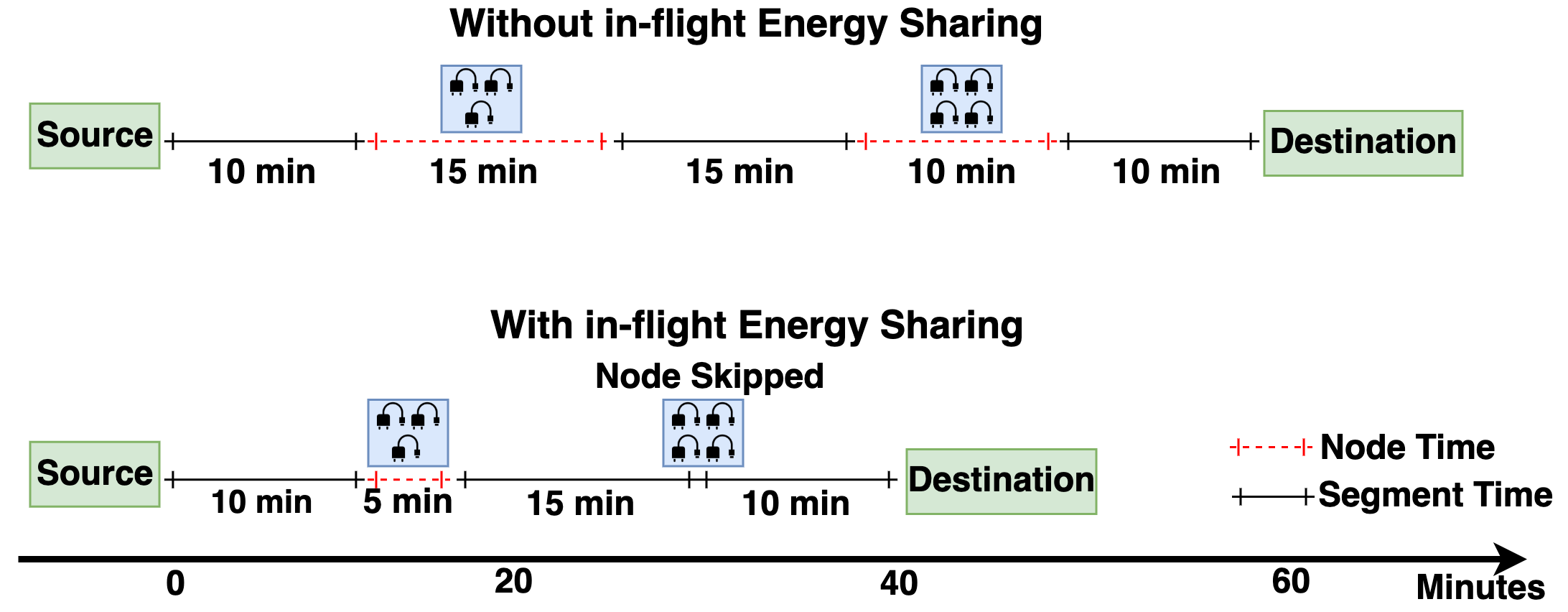}
  \caption{Effect of In-Flight Recharging on Swarm-based Drone Services Composition}
  \label{fig:effect}
\end{figure}

Consider the scenario of delivering medical supplies using a swarm of drones in a city. We consider a skyway network where the nodes represent buildings rooftops. Each rooftop may be a delivery destination, a recharging node with a number of stations or both (Fig.  \ref{fig:motivating}). Let us assume that a swarm is used to deliver multiple packages from supplier to a hospital. The requirement is that all packages must arrive as soon as possible and together within a predefined timeframe \cite{wang2009web}. Such requirement stem from the inherent interdependencies of the supplied packages. For example, certain medical drugs need to be mixed at the hospital but compounds would need to be delivered separately and within a timeframe \cite{tripathi2013essentials}. Hence, they need to arrive at the same time. Fig. \ref{fig:motivating} represents a skyway network. We assume that the packages requested are of different weights. An optimal composition of swarm-based drone services results in the optimal path from source to destination with the fastest delivery time operating under a set of constraints. \textcolor{black}{The constraints include different intrinsic and extrinsic constraints. Intrinsic constraints include the different energy consumption rates in the swarm due to different payloads and the battery capacities. A drone carrying a heavier payload typically consumes more energy than a drone with lighter payload \cite{alkouz2020swarm}. The extrinsic constraints include the limited recharging pads at a station, wind conditions, and the time-constrained arrival of packages.} We assume that swarms are \textit{static}: 1. the number of drones in a swarm are constant, and 2. drones in a swarm stay together spatially until they reach their destination \cite{akram2017security}. We also assume that the optimal swarm \emph{formation is preset, i.e., has already been decided and selected,} based on the averaged wind conditions and will not change until packages are delivered \cite{alkouz2020formation}. \textcolor{black}{ We adopt the five formations proposed in \cite{alkouz2020formation}, including Column, Front, Echelon, Vee, and Diamond. These formations are popular formations inspired by bird flocks and military airforce for energy conservation purposes and protection \cite{alkouz2020formation}.} An optimal formation is one that consumes the least amount of energy under certain wind conditions \cite{alkouz2020formation}. We assume that each skyway segment may have a different wind speed and direction.
We aim to minimize the swarm-based delivery time by \emph{reducing the number of stops and recharging time at intermediate nodes in the skyway network } (Fig. \ref{fig:effect}). The recharging time of a swarm at an intermediate node includes time for recharging and waiting. A drone must wait to be recharged where there is no available recharging station. \textcolor{black}{We assume that all the recharging pads are similar and hence the charging rate is same.} In this context, given a preset swarm formation, we propose leveraging the concept of in-flight charging to reduce the number of stops and recharging time at intermediate nodes. We aim to adopt the concept of \textcolor{black}{Energy-as-a-Service} and implement it within a swarm. As shown in Fig. \ref{fig:motivating}, a swarm has a designated \emph{support drone} whose sole job is to recharge other drones in the swarm. We assume that the support drone carries a larger battery to share energy with other drones. In that respect, the support drone is considered as an Energy-as-a-Service (EaaS) provider. The rest of the drones are EaaS consumers. The energy sharing is assumed to occur during the swarm flight in a segment. Since there are multiple EaaS consumers and a limited flight time within a segment, there is a need for an EaaS provider (i.e, support drone) to select an optimal set of energy requests emanating from EaaS consumers over a period of time. We assume that the provider can share energy only with a single consumer at a time. Hence, the provider should select the best consumer candidate at a time.\looseness=-1

In most types of formations (e.g., Vee formation), each drone would consume a different amount of energy which is dependent on its position in the formation due to wind drag forces and upwash/downwash forces generated by front drones \cite{alkouz2020formation}. The in-flight energy sharing occurs at less than 1.2 meters distance between the provider and consumer \cite{benbuk2020leveraging}. We assume that the position of the support drone should be neighboring the currently selected consumer. Therefore, the swarm may need to reorder itself to facilitate energy sharing. Additionally, the reordering problem is compounded by the different energy consumption rates of the drones according to their positions. In that respect, \emph{we propose to study the position of a support drone in a swarm} to examine the effect of reordering the swarm on energy consumption.

\section{Related Work}
We identify there areas of research which are related to our work: \emph{(1) swarm-based drone services, (2) energy sharing and wireless energy transfer, and (3) service selection and composition}. We describe the related work in each of these areas.

\subsection{Swarm-Based Drone Services}
%Drones in Delivery
%challenges include energy

The use of drones for delivery is receiving increasing investment and interest from industry and research, respectively. Drones are used to deliver medicine \cite{scott2017drone}, post \cite{bamburry2015drones}, and pizza \cite{bamburry2015drones}. The benefit of using drones in delivery over traditional methods is time sensitivity \cite{scott2017drone}. Drones are fast and can deliver packages to areas that are hard-to-reach with vehicle based services \cite{lin2018drone}. A natural progression of using single drones in different \textcolor{black}{applications, including} delivery, is the use of multiple drones cooperatively to achieve a common goal.\looseness=-1

%Recent literature explores the augmentation of single drones capabilities in delivery and other applications using drone swarms.

%Drone swarms applications
%one of the examples is delivery
Tackling tasks as a collective unit adds to the benefits of a single drone. \emph{Drone swarms} could cut down the time needed in target search \cite{cimino2015combining},  surveillance missions \cite{sharma2018lorawan}, site mapping \cite{casella2017mapping}, and agriculture applications \cite{mogili2018review}. In addition, it could achieve tasks that a single drone is not capable of doing. This includes fascinating sky shows \cite{waibel2017drone} and the delivery of multiple packages to the same destination at the same time \cite{alkouz2021reinforcemnt}. Swarm-based drone delivery is an emerging application that has been recently explored in the literature \cite{alkouz2020swarm}. Multiple works in swarm-based deliveries refer to coordinating multiple individual operated drones as drone swarm \cite{san2016delivery} \cite{kuru2019analysis}. However, \emph{we refer to a swarm as a set of drones that act as a single entity and are bounded by a space window.} 

%removed after review
% The large-scale adoption of drones in deliveries come with technical challenges to overcome. These challenges include the development of quieter drones \cite{funk2018human}, collision avoidance \cite{murray2015flying}, possible flight paths \cite{dorling2016vehicle}, and limited flight times \cite{shahzaad2019composing}. The last is mainly concerned about the energy consumption and the battery capacities of drones. A drone's average flight time is about 30 minutes \cite{lee2015autonomous}. Hence, solutions to overcome the energy limitations in single drone deliveries are proposed. 

%Drone energy solutions
%solved by formations
The large-scale adoption of drones in deliveries come with technical challenges to overcome. Power is considered the bottleneck of the drone industry \cite{long2018energy}. This item is of particular interest when travelling long distances to deliver packages or medicine \cite{scott2017drone}. Solutions proposed include creating new types of batteries that live longer than the traditional Li-Po batteries \cite{kardasz2016drones}. Solar powered drones are also developed to extend the battery life \cite{d2016suav}. CyPhy proposed a micro-filament solution to have an infinite flying time by connecting the drone with a micro-filament wire to batteries on ground \cite{quick_2015}. Battery replacements throughout travel and the use of wireless charging stations midway is also proposed \cite{hamdi2021drone}. However, these solutions suffer of lack of mobility, weather condition reliance, human-power dependability, and added recharging times. In drone swarms, an alternative solution is proposed that utilizes \emph{drone flight formations} to enhance the energy consumption during delivery missions \cite{alkouz2020formation}.\looseness=-1

%https://www.youtube.com/watch?v=6wk9RZ8kUX8
%https://www.commercialuavnews.com/infrastructure/power-bottleneck-drone-industry

%Drone swarm formations and energy
%further enhancement we propose energy sharing

The study of flight formations have been widely explored in the fields of military and birds migrations. A swarm in military takes different formations to allow for better protection, view, and aiming \cite{haissig2004military}. Birds take different formations for energy conservation purposes \cite{cutts1994energy}. As a bird flaps, a rotating vortex of air rolls of at each of its wingtips. This causes the air directly behind the bird to be pushed downwards (downwash). In contrast, the air to the side of the bird gets pushed upward (upwash). The upwash forces give a free lift to the birds at the back causing them to consume less energy on flapping. Hence, we see birds flying in Vee formations for instance \cite{cutts1994energy}. The heart rate of birds flying in a vee formation were found to be less indicating less energy consumption \cite{weimerskirch2001energy}. This is specially important when birds travel for long distances during migration.
%This is caused by the trailing forces generated by the front birds in the form of upwash and downwash forces.

In swarm-based drone applications, the study of formation flying is still in its infancy. A swarm of UAV's fly in different geometrical flight formations for 3d scene reconstruction purposes \cite{milani2016impact}. For energy conservation purposes, using a Vee formation in a fixed wing UAV swarm has proven to consume less energy and behave similar to birds \cite{mirzaeinia2019energy}. In addition, the effect of different formation flying on energy consumption of quadcopters was studied under different wind conditions \cite{alkouz2020formation}. Flying in a formation proved to consume less energy for different wind conditions due to reduced drag and upwash/downwash forces. The study also presented that drones consume different amounts of energy based on their \textit{position} in a formation \cite{alkouz2020formation}. However, to the best of \textcolor{black}{our} knowledge, the effect of reordering the drones in a formation was not explored previously. We propose to optimise the distribution of energy consumption between the drones in the swarm by \emph{re-ordering the drones within a formation}. In addition, we propose to overcome the power limitation in swarm-based deliveries using \emph{in-flight energy sharing} within a swarm.

%Removed after review
%The reduced drag forces in a formation is a phenomenon seen in auto racing cars as well. The behavior is known as drafting. In this technique the cars drive close to each other which reduces the aerodynamic drag force that can slow them down. In this way, they conserve fuel and obtain greater speeds as well \cite{romberg1971aerodynamics}.

\subsection{Energy Sharing and Wireless Energy Transfer}
%Amani/Abdallah
%energy sharing
%energy sharing in different applications
%how it works
%how are we different
%Sharing energy enables the development of self-sustaining ecosystems.
There is a growing interest in wireless energy charging in the field of wireless sensor networks, internet of things, mobile social networks, vehicular ad hoc networks, and UAV networks \cite{lakhdari2020Vision}. A new paradigm of radio wave-based uncoupled wireless charging has enabled sharing and accessing  harvested energy from IoT devices \cite{bell2019systems}.   Multiple wireless charging methods were introduced in IoT and wireless sensor networks \cite{lakhdari2020Vision}.  For example, a reliable energy supply method was proposed to charge low battery IoT devices using a mobile charger in the network  \cite{na2017energy}.% Another approach proposed the use of drones  to  wirelessly charge rechargeable sensor networks \cite{chen2016mobile}. %Another model was proposed to charge light-weight drones entirely through wireless means. % I can't find the refernce yet

Wireless energy sharing services have been introduced recently as an alternative ubiquitous solution to charge IoT devices \cite{lakhdari2020Vision}\cite{yao2022wireless}. Energy service is defined as the wireless energy transfer among IoT devices \cite{lakhdari2020composing}. Several methods were proposed to charge IoT devices using energy services \cite{lakhdari2020Vision}\cite{Amani2022QoE}\cite{lakhdari2021fairness}. A temporal composition algorithm was proposed to charge IoT devices in confined areas \cite{lakhdari2020composing}.  Another approach was proposed to charge IoT devices from highly fluctuating IoT energy providers \cite{lakhdari2020Elastic}. Mobility pattern impact on IoT energy services was addressed in \cite{lakhdari2020fluid}.  Other energy sharing compositions were proposed to address charging mobile IoT devices  \cite{abusafia2020Reliability}\cite{lakhdari2021proactive} \cite{lakhdari2020fluid}. To the best of our knowledge, none of the previous work studied the use of drones as wireless chargers to reduce the delivery time in drone delivery services. \looseness=-1

\subsection{Services Selection and Composition}
Service is the highest level in the computing chain where actions are built on knowledge \cite{bouguettaya2017service}. A service consists of functional and non-functional (Quality of Service) aspects \cite{yu2012multi}. The key differentiator between services are the QoS measures. In this paper, we propose to leverage the service paradigm to model the functional and non-functional (QoS) properties to compose swarm-based drone delivery services and energy sharing services. Hence, we look at the work done in service composition.

The challenges in service computing can be grouped into four areas: service design, service composition, crowdsourcing-based reputation, and IoT \cite{bouguettaya2017service}. The work proposed in this paper lies mainly under the two umbrellas of service composition and IoT. Drones play a role in IoT as they are dependent on sensors and embedded software to provide communication to achieve their goal \cite{arasteh2016iot}. In drone services specifically, a model for single drone-based services was proposed \cite{hamdi2021drone}. The single drone-based services model was defined by its functional and non-functional (QoS) properties including delivery time and cost. Each skyway segment is abstracted as a drone service which can be served by a single drone. Their work only focuses on the composition of single drone services. This concept was augmented to swarm-based drone service to compose the optimal path of drones swarm from a source to a destination under various constraints \cite{alkouz2020swarm}. The concept of wireless energy sharing services between IoT devices has been previously defined as Energy-as-a-Service ($EaaS$) \cite{lakhdari2020composing}. The existing work focused either on composing services in a confined area to fulfil the consumers queries \cite{lakhdari2020composing} \cite{lakhdari2020Elastic} or composing energy requests to maximize the utilization of spare energy \cite{abusafia2020incentive}\cite{abusafia2020Reliability}. To the best of our knowledge, there is no previous work on composing swarm-based delivery services based on swarm formations, energy sharing, and re-ordering. Hence, this paper is the first attempt to model energy and time constrained delivery environment for swarm-based drone deliveries using \textit{in-flight energy sharing} and formation re-ordering using a service-oriented approach.

%To be changed if more details in EaaS
\section{System Architecture and Composition Model}
\label{architectureandmodel}
%Do we need architecture? i.e. SOA registries/interfaces/edge/cloud/etc.

\begin{figure}[!t]
\centering
  \centering
  \includegraphics[width=0.8\linewidth]{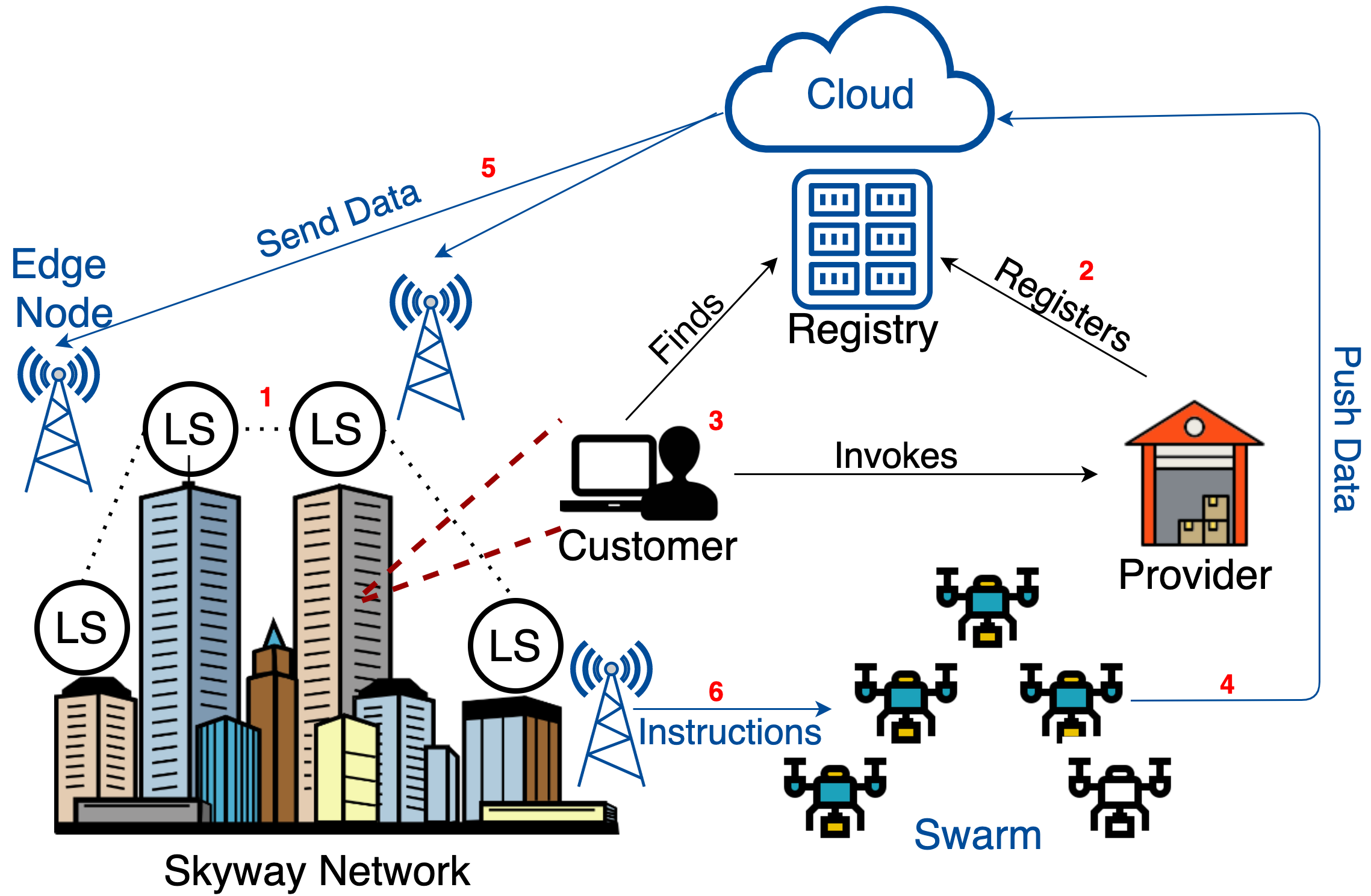}
  \caption{High Level Swarm-based Drone Services System Architecture}
  \label{fig:architecture}
\end{figure}

% We present a high level system architecture for swarm-based delivery services. As shown in the \textcolor{red}{numbered} figure \ref{fig:architecture}, the architecture is premised on the delivery of multiple packages to a single destination in a line-of-sight skyway network with landing spots (LS) at each node (rooftops) (\textcolor{red}{1}). The architecture is an adapted service oriented architecture (SOA) with customers, providers (warehouses), and registries. The \emph{provider} registers its services in the registry to advertise them (\textcolor{red}{2}). The \emph{customers} would view and select the desired services from the registry and invoke them (\textcolor{red}{3}). When a customer invokes an SDaaS service, the delivery management system determines the set of drones needed, including the support drones, and the path to be taken. The delivery management system shares the computations among the swarm, cloud, and edge nodes.
%====================================================================================================
We present a high-level system architecture for swarm-based delivery services. As shown in the numbered \textcolor{black}{Fig. \ref{fig:architecture}}, the architecture is premised on the delivery of multiple packages to a single destination in a line-of-sight skyway network with landing spots (LS) at each node (rooftops) (\textcolor{black}{1}). The architecture is an \textcolor{black}{adopted} service-oriented architecture (SOA) with consumers, providers (warehouses), and registries. The \emph{provider} registers their swarm-based delivery services in the registry for advertisement (\textcolor{black}{2}). \emph{Consumers} would view, select, and invoke the desired services from the registry (\textcolor{black}{3}). When a customer invokes a swarm-based drone service, the delivery management system determines the set of drones needed, including the support drones and the path to be taken. The delivery management system shares the computations among the swarm, cloud, and edge nodes.

\textcolor{black}{Simple computations including analysis of sensor data and battery information will be computed at the \emph{swarm} level. More \textcolor{black}{sophisticated and expensive computations} will be performed at the \emph{edge} level. These computations include swarm-based drone services and EaaS compositions. The edge nodes are distributed in strategic locations to aid delivery. It allows for lower latency and faster communication speed between the swarm and the edge. Computations requiring lots of data will be done on the \emph{cloud} level.} Such computations include navigation throughout the skyway network. As the swarm moves, the drones communicate their data, i.e., battery states and locations, to the cloud for storage purposes (\textcolor{black}{4}). The cloud advertises this data to the edge nodes at frequent intervals to help in decision making (\textcolor{black}{5}). The swarm receives multiple instructions such as picking up the packages, path to the destination, recharging, etc. these instructions are communicated with the swarm from the cloud through the edge server (\textcolor{black}{6}). On arrival of the swarm to the destination, the customer is notified.

We abstract each swarm traveling in a  skyway segment as a swarm-based drone service, as explained earlier. Our goal is to select and compose the optimal set of swarm-based drone services from a source point to a destination point given the highly constrained environment due to different carried payloads, limited charging spots, and constrained delivery time window. We then aim to further optimize the swarm-based drone service composition by incorporating \textcolor{black}{Energy-as-a-Service (EaaS)}. This paper considers the environment to be deterministic, i.e., the surrounding conditions do not change after compositions.

%SDaaS formal definition
% needs rephrasing and changes
We formally define a Swarm-based Drone Service. Then we define a swarm-based drone service customer request. Later, we formally define an Energy-as-a-Service (EaaS).\\
\textbf{Definition 1: Swarm-based Drone Service.} It is defined as a tuple of $<SDS\_id, S, F>$, where
\begin{itemize}
    \item $SDS\_id$ is a unique identifier.
    \item $S$ is the swarm travelling in a swarm-based drone service. $S$ consists of: 
        \textcolor{black}{\begin{itemize}
            \item  $D$ which is the set of delivery drones and support drones forming $S$, a tuple of $D$ is presented as $<D_{d1}, D_{d2}, .., D_{dn}>$ and $<D_{s1}, D_{s2}, .., D_{sm}>$.
            \item The battery levels of every $d$ in $D$ $<b_1, b_2, .., b_n>$.
            \item The payloads every $d$ in $D$ is carrying $<p_1, p_2, .., p_n>$.
        \end{itemize}}
     
    %\hl{add that D includes delivery drones and support drones}
    \item \textcolor{black}{$F_{SDS}$} describes the delivery function of a swarm on a skyway segment between two nodes, A and B. F consists of the travel time $tt$,  charging time $ct$, and waiting time $wt$ when recharging pads are not enough to serve $D$ simultaneously in node B. \textcolor{black}{In many instances, the number of delivery drones $D$ would be larger than the number of available recharging pads at a node. Therefore, the drones will not be able to recharge simultaneously and will have to wait for each other.}
\end{itemize}

\textbf{Definition 2: Swarm-based Drone Service Request.} A request is a tuple of $<\alpha, \beta, P>$. $\alpha$ is the source node, $\beta$ is the destination node, and $P$ are the weights of the packages requested, where $P$ is $<p_1,p_2, ..p_n>$.

\textbf{Definition 3:  In-flight Energy-as-a-Service (EaaS).} We adopt the definition of $EaaS$ in\cite{lakhdari2020composing}.  An $EaaS$ is defined as a tuple of $<id, F,Q>$, where:
\begin{itemize}
    \item ${EaaS_{id}}$ is a unique service identifier.
    \item \textcolor{black}{$F_{EaaS}$} is the function of sharing energy by a support drone. 
    \item $Q$ is a tuple $<ae, loc,st, et>$ where each attribute donates a $QoS$ property of $EaaS$ as following: $ae$ is the amount of available energy that a support drone can share, $loc$ is the location of a support drone $<x,y>$, $st$ is the start time of a support drone's $EaaS$, and $et$ is the end time of a support drone' $EaaS$.
\end{itemize}

\textbf{Definition 4: In-flight Energy Service Request (ER).} We adopt the definition of $ER$ in \cite{abusafia2020incentive}. An $ER$ request is defined as a set of $<{id}, F, QR>$, where:
\begin{itemize}
   \item ${ER_{id}}$ is a unique energy service request identifier.
   % \item $D\_{id}$ is a unique delivery drone's identifier
    \item \textcolor{black}{$F_{ER}$} is the function of receiving energy by a delivery drone.
   \item $QR$ is a tuple $<re,loc, st, et>$ where each attribute donates a requirement property of $ER$ as following: $re$ is the amount of requested energy by a delivery drone, $st$ and $et$ are the start and end times of the delivery drone's receiving duration, and $loc$ is the location of the delivery drone. 
\end{itemize}

\textcolor{black}{Table \ref{tab:symbols} serves as a reference for the meanings of the main symbols used in this paper.}

\begin{table}[ht]
\caption{\textcolor{black}{Summary of Notations}}
\label{tab:symbols}
\begin{tabular}{l|l}
\hline
Symbol & Meaning                                                                                                                                                    \\ \hline
$SDS$    & Swarm-based Drone Service                                                                                                                                  \\
$S$      & Swarm                                                                                                                                                      \\
$D_d$    & Delivery drone                                                                                                                                             \\
$D_s$    & Support drone                                                                                                                                              \\
$dt$     & Total delivery time of the packages                                                                                                                        \\
$tt$     & Travel time within a segment                                                                                                                               \\
$nt$     & Node time (time spent at a node to recharge the swarm)                                                                                                     \\
$ct$     & \begin{tabular}[c]{@{}l@{}}Charging time (time spent by a drone recharging, \\ depends on the charging pad rate and required energy)\end{tabular}            \\
$wt$     & \begin{tabular}[c]{@{}l@{}}Waiting time (time spent by drones waiting for other drones to \\ finish recharging due to limited number of pads)\end{tabular} \\
$EaaS$   & Energy-as-a-Service                                                                                                                                        \\
$ER$     & Energy request                                                                                                                                             \\
$st, et$ & Start and End times of a provisioned energy request                                                                                                        \\ \hline
\end{tabular}
\end{table}

\subsection{\textcolor{black}{Problem Formulation}}
\textcolor{black}{Given a swarm-based drone service request from a consumer, the problem is formulated as composing the best $SDS$ services from the source node $\alpha$ to the destination node $\beta$ using in-flight energy sharing. The composition is constrained by several constraints. First, the payload and battery limitations of the drones are intrinsic constraints. Second, the availability of the charging pads and the wind conditions are extrinsic constraints. Third, the rate of in-flight recharging might not be sufficient, in some instances, to cover the energy loss due to formation reordering and re-positioning the drones within a formation. As described earlier, drones consume different amounts of energy based on their positions within a formation. Moreover, there are a set of challenges when addressing $SDS$ composition with in-flight energy sharing. First, the selection of which delivery drone to charge, when to charge, and for how long to charge affect the planning of the journey. Second, the composition of the next service, in addition to the aforementioned constraints, depends on the gain provided with in-flight energy sharing. Third, deciding the number of support drones $D_s$ to carry out the mission is an extra challenge that depends on various number of variables like the size of the swarm, the carried payloads and the wind conditions. }

\textcolor{black}{Given all the aforementioned constraints and challenges, our goal is to optimally compose two level services. First, composing $SDS$ services from the source to the destination with the shortest delivery time ($\min(dt)$). Second, composing $EaaS$ services within every $SDS$ service. In the second level $EaaS$ composition, the goal is to allocate the energy needing delivery drones to the support drones for an overall energy gain that will reduce the charging time at the next node. The $EaaS$ provisioning is constrained by the start times $st$ and end times $et$ of energy requests $ER$ by the delivery drones $D_d$ and the $st$ and $et$ of the support drones $EaaS$ service. }

%Types of compositions
    % Deterministic
    % Stochastic

\section{Swarm-based Drone Services Composition Framework}
\begin{figure*}
    \centering
    \includegraphics[width=0.9\linewidth]{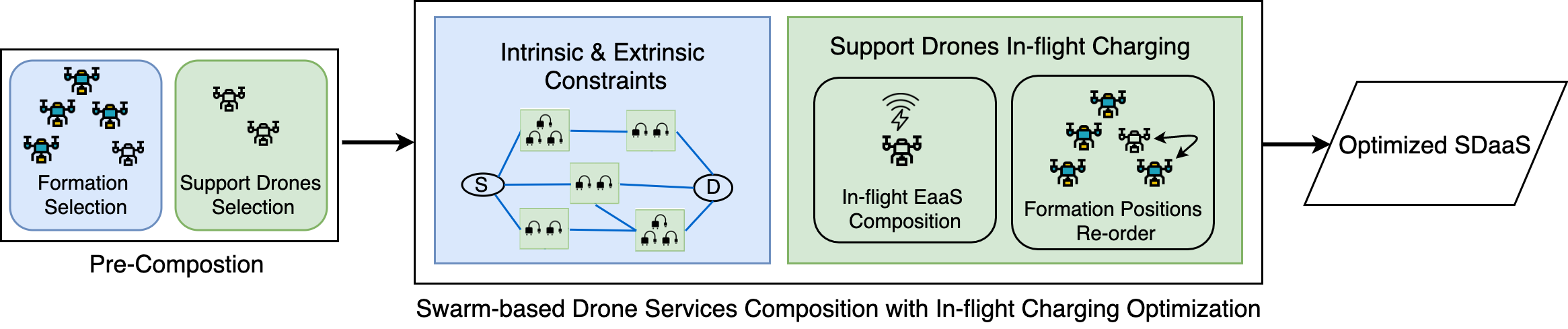}
    \caption{Swarm-Based Drone Service Composition Framework}
    \label{fig:Composition_framework}
\end{figure*}

The composition framework consists of two main modules namely, swarm-based drone services \textit{Pre-Composition}, and swarm-based drone services \textit{Composition with in-flight charging}. The second module is a nested two-level swarm-based drone service $SDS$ and $EaaS$ compositions. The result of these modules is an optimized composition of swarm-based drone services $SDS$. Fig. \ref{fig:Composition_framework} depicts the steps involved in the proposed $SDS$ composition framework.
 
\subsection{Swarm-based Drone Service Pre-Composition}
Before composing the set of optimal services from a source to a destination we need to select the optimal swarm formation and support drones. In this section, we discuss how to select the optimal formation as proposed by \cite{alkouz2020formation}, select the optimal number of support drones, and where to position the support drones in a formation.

\subsubsection{Formation Selection}
\begin{figure}[!t]
\centering
  \centering
  \includegraphics[width=0.6\linewidth]{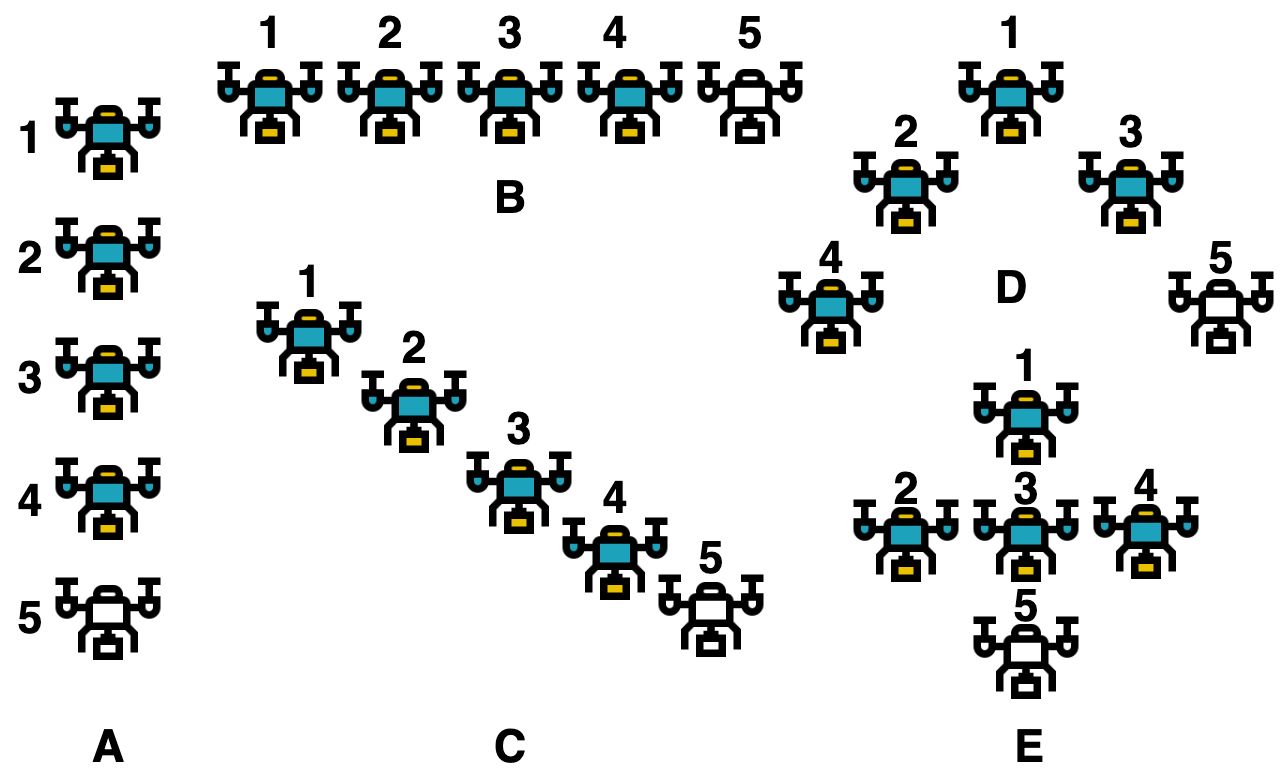}
  \caption{Different Swarm Formations A:Column, B:Front, C:Echelon, D:Vee, E:Diamond}
  \label{fig:swarmFormation}
\end{figure}
% The selection of the swarm formation is essential for an energy optimized SDaaS services composition. Drones in different swarm formations consume different amount of energy due to two reasons. First, a drone in formation experience upwash and downwash forces from other drones in the swarm. These forces cause drones to consume different amount of energy based on their position in the formation. We assume that each two neighboring drones in a formation are 1m apart. Second, a drone in a formation experiences different amount of drag forces based on their position and the external wind direction and speed. Hence, to select the optimal formation, we assume that a \emph{formation is fixed}. This means that once an optimal formation is selected initially, it stays fixed throughout the journey to the destination. The optimal formation is selected based on averaged wind conditions in the delivery environment. Looking at all the SDaaS services in the skyway network, we compute the average wind speed of all SDaaS and find the most frequent wind direction and select them as the averaged wind conditions. We identify five formations including Column, Front, Echelon, Vee, and Diamond. Figure \ref{fig:swarmFormation} shows the different formations. We adopt the study done in \cite{alkouz2020formation} to compute how much each drone consumes energy in different formations under different wind conditions.
%==============================================================================================================

The selection of the swarm formation is essential for an energy-optimized composition of $SDS$ services \cite{alkouz2020formation}. Drones in different swarm formations consume different amounts of energy due to two reasons. First, a drone in a formation experiences upwash and downwash forces from other drones in the swarm. These forces cause drones to consume different amounts of energy based on their position in the formation. %We assume that every two neighboring drones in a formation are one meter apart. 
 Second, a drone in a formation experiences diverse drag forces based on its position and the external wind direction and speed. Hence, \textcolor{black}{it is crucial to select an optimal formation to minimize the energy consumption of the $SDS$. For example, with a front wind, drones in a vee formation consume the least amount of energy compared to other formations. On the other hand, with a side wind, a diamond formation consumes less energy \cite{alkouz2020formation}. In this paper, we adopt a \emph{fixed formation} for the swarm-based drone services \cite{alkouz2020formation}; once an optimal formation is selected initially, we assume that it will not change throughout the journey to destination.} Five formations are identified, including Column, Front, Echelon, Vee, and Diamond. Fig. \ref{fig:swarmFormation} shows the different formations. We adopt the study done in \cite{alkouz2020formation} to compute how much each drone consumes energy in different formations under different wind conditions.

%The optimal formation is selected based on averaged wind conditions in the delivery environment. Looking at all the swarm-based drone services in the skyway network, we compute the average wind speed of all swarm-based drone services and find the most frequent wind direction and select them as the averaged wind conditions. 

% \hl{do I have to put the results table of every drones energy consumption in different formation? }

\subsubsection{Support Drones Selection}

\textcolor{black}{Redundancy is the duplication of critical components of a system to either increase the reliability of a system or to improve actual performance \cite{shooman2002reliability}. In this paper, we adopt the concept of redundancy to improve the performance in terms of energy consumption and delivery time by having \emph{support drones}. We consider the inclusion of the support drones to be equivalent to redundancy. This is because we assume the support drones have similar specifications to the delivery drones. The only difference is the payload carried that comprises of extra batteries and the drones are equipped with energy transmitters. This is analogous to disk drives, which can be redundant for efficiency rather than just reliability \cite{patterson1988case}.} In this step, we study the optimal number of support drones for a particular delivery request using the redundancy theory \cite{shooman2002reliability}. We then study the options for positioning the support drones in the formation before the $SDS$ services composition. 

There are different types of redundancies based on the needs of a system \cite{shooman2002reliability}. These include $N+1$, $N+2$, $2N$, and $2(N+1)$ redundancies, where $N$ is the number of components in the system. In our case, $N$ is the number of the delivery drones in the swarm before adding the support drones. Each redundant component, i.e., support drone, added to the system decreases the probability of failures. In our scenario, a failure (i.e., unsuccessful delivery) occurs when a swarm cannot reach the destination due to the battery limits of drones. In this paper, we assume that the drones in the swarm are of the same size and specifications. For example, we use DJI Phantom 3 drone model for all the experiments in this paper. We also assume that the max payload a drone can carry is 1.4kg. The battery that the DJI Phantom 3 uses weights 365g. Hence, in the case of this drone model, we assume a support drone can carry three additional batteries increasing its energy capacity. We also assume that all the drones are initially fully charged.
%========================================================================================================================================================
We consider the following attributes that affect the delivery to compute the probability of failure and select the optimal number of redundant components, i.e., support drones, for a certain delivery request:
\begin{itemize}
    \item The size of the swarm without the support drones $S_s$. This is based on the number of packages in an $SDS$ request as explained in section \ref{architectureandmodel}. 
    \item The payload ratio for each drone $D_{pr}$. The payload directly affects the energy consumption in a drone. Hence, we compute the average payload ratio for all the drones in the swarm using the following equation:
    \begin{equation}
        D_{pr} = \frac{\sum_{i}^{n}\frac{p_{i}}{max~payload~ capacity}}{no.~of~drones}
    \end{equation}
    where $p_i$ is the payload drone $i$ is carrying, the max payload capacity is 5kg, and the no. of drones is the total number of drones in the swarm before adding any support drones.
    \item Distance between the source and the destination $d(\alpha,\beta)$. We use  Dijkstra's shortest path to compute the distance.
    \item Energy capacity of a support drone $E_c$. This is 3+1=4 times of the delivery drone including the original battery as explained earlier.
    \item The averaged wind conditions $W$. As described earlier, wind highly affects the energy consumption of a swarm. Hence, the wind factor needs to be considered in support drones selection.
\end{itemize}

We use the aforementioned delivery affecting factors to compute how many redundant components, i.e. support drones, are needed. We do this by computing the probability of failure and matching the probability with the equivalent redundancy values in Table \ref{tab:redundancy}. We compute the probability of failure using the following equation:

\begin{equation}
     P = \prod_{i=1}^{4}w_i * \widetilde{p_i}
\end{equation}

where $p_{1}=d_{pr}$ is the drone payload ratio, $p_{2}= d(\alpha,\beta)$ is the distance between the source and the destination, $p_{3}= E_{cp}$ is the energy capacity of the support drones, and $p_{4}=W$ is the averaged wind conditions. The weights $w_i$ in the equation are computed based on how much each factor \emph{consumes energy per unit time independently}. For example, let us assume that a swarm consumes 3\% of its total power to travel for 1 minute at a speed of 30 km/hr, hence, $w_2 = 0.03$. The distance between source and destination $d(\alpha,\beta)$ is 5km. Therefore, $P_{i=2} = 0.03 * ((5/30)*60) = 0.3 $. The same is done to compute the rest of the weights $w_i$. Instead of taking the true value of each factor, we normalize them between 0 and 1 to unify the different ranges of the affecting factors $\widetilde{p_i}$. Once the probability of failure is computed, the number of redundant components (support drones) are decided according to Table \ref{tab:redundancy}.

%\hl{find the exact weights and add them}

\begin{table}[ht]
\caption{Redundancy Values for Different Failure Probabilities}
\label{tab:redundancy}
\centering
\resizebox{0.8\linewidth}{!}{%
\begin{tabular}{|l||l|}
\hline
Failure Probability P & Redundancy Number \\ \hline
0-19                  & N+1               \\
20-39                 & N+2               \\
40-59                 & N+3               \\
60-79                 & N+4               \\
80-100                & 2N                \\ \hline
\end{tabular}%
}
\end{table}

\subsubsection{Support Drones Positioning} 

\begin{figure}[!t]
\centering
  \centering
  \includegraphics[width=0.5\linewidth]{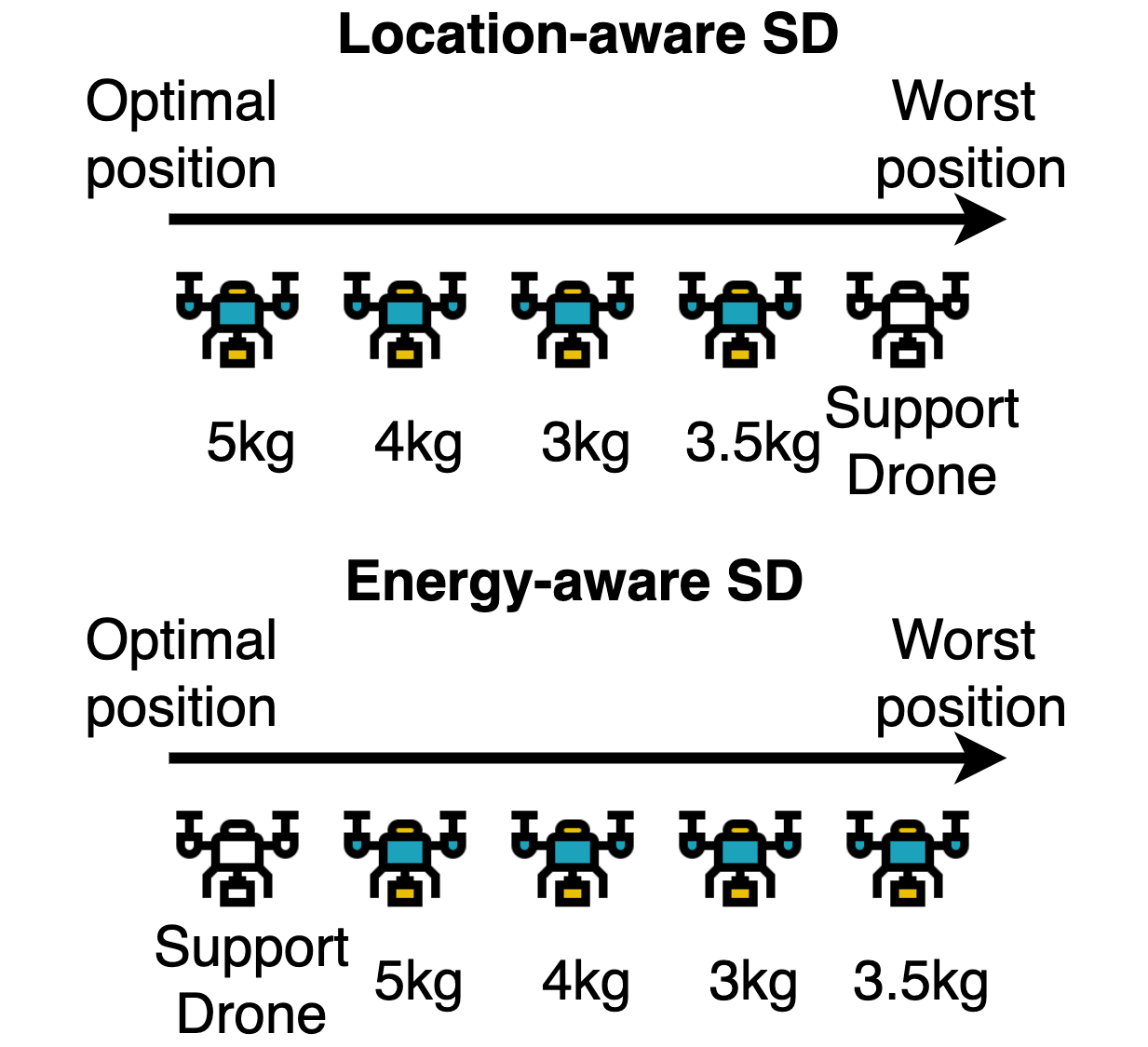}
  \caption{Support Drones Positioning}
  \label{fig:initalpositoining}
\end{figure}

Once the number of support drones in a swarm is decided, we need to study the optimal position to place the support drones in the swarm. The positioning is important for a couple of reasons. First, a support drone can charge other drones up to 1.2 meters distance \cite{benbuk2020leveraging}. Second, the position of the support drones in the swarm affects their energy consumption due to upwash/downwash and drag forces in different formations. The drones in different formations consume energy differently based on their positions in a formation and the wind direction. For example, a drone at the front of a Vee formation consumes the most energy if the wind is coming from the front \cite{alkouz2020formation}. In this paper, we propose to study the effect of two positioning settings. Fig. \ref{fig:initalpositoining} describes the two settings. Note, in this step, \textit{we are only selecting the initial position of a support drone}. Later, in \textcolor{black}{section} \ref{reordering}, we will re-order them as needed.

\begin{itemize}[leftmargin=*]
    \item Location-aware Support Drone: Looking at the full swarm, i.e. the delivery drones and the support drones, we position the delivery drone carrying the maximum payload at the optimal position in the formation. Descendingly, we position the drones carrying the heavier payloads in the optimal positions and the least payload carrying in the worst positions. Lastly, we position the support drones in the worst positions. This setting would ensure that the delivery drones are consuming the least amount of energy due to the positioning. However, it comes with the cost of non-optimal support drones positioning. The support drones are aware of their location and \textcolor{black}{give} up their prime position for the sake of optimal positioning of the delivery drones.
    
    \item Energy-aware Support Drone: Looking at the full swarm, i.e. the delivery drones and the support drones, we position the support drones at the optimal position in the formation. Then descendingly, as the first setting, we position the delivery drones from the optimal positions to the worst positions based on their carried payloads. This approach would ensure that the support drones are consuming the least amount of energy due to their positions and would have more energy to share with the delivery drones. However, the delivery drones would be consuming more energy in this approach. The support drones are aware of their energy and are positioned in the least energy consuming positions.
\end{itemize}
As can be seen from the two settings, we need to compromise between the optimal positioning of the delivery drones and the optimal positioning of the support drones. Hence, in section \ref{expirementsandresults}, we run experiments to test both approaches and reach a conclusion of which setting to be adopted.

%Given m support drones where to position them?
    % fixed position
    % flexible position
    %what each type means and how it affect the consumption and composition of energy services

\subsection{Swarm-based Drone Services Composition with In-Flight Energy Sharing}
\label{composition}
%SDaaS  algorithms
% After pre-composition, we select and compose the optimal SDaaS services from the source to the destination using a modified A* algorithm. Optimality, in this paper, is a single objective of composing SDaaS services with the least delivery time. We assume the swarm is \emph{static}, i.e. the members are decided at the source node and no additions or reduction of members occurs throughout the path to the destination \cite{alkouz2020swarm}. Given a connected network, we compose the best available path from the source to the destination. The composition of the optimal services mainly  depend on two components as shown in Fig. \ref{fig:Composition_framework}. The first are the intrinsic and extrinsic constraints surrounding the delivery environment. The second is the in-flight energy sharing process that will affect the composition of the services.

%=========================================================================================================================================
After pre-composition, we select and compose the optimal swarm-based drone services from the source to the destination using a modified A* algorithm. Optimality, in this paper, is a single objective of composing $SDS$ services with the least delivery time. We assume the swarm is \emph{static}, i.e. the members are decided at the source node and no additions or reduction of members occurs throughout the path to the destination \cite{alkouz2020swarm}. Given a connected network, we compose the best available path from the source to the destination. The composition of the optimal services mainly  depend on two components as shown in Fig. \ref{fig:Composition_framework}. The first is the intrinsic and extrinsic constraints surrounding the delivery environment. The second is the in-flight energy sharing process that will affect the composition of the services.

%the intrinsic and extrinsic
A swarm-based delivery environment is highly affected by multiple constraints. The intrinsic constraints include the different energy consumption rates between the drones due to \emph{payload} and \emph{position in a formation}. The extrinsic constraints include the \emph{wind conditions} that highly affect the energy consumption of a swarm and the different number of \emph{recharging stations} available at every node to serve the swarm. The proposed swarm-based drone services composition framework addresses the aforementioned constraints and optimizes the selection of the services through in-flight energy sharing. \looseness=-1

%the in-flight energy sharing process

%the composition method and algorithm

%At this step, we assume that the support drones are not yet effective. They will be used to further optimize the composition in subsection \ref{furtheroptimization}. 

%The composition here solely depends on the capability of the delivery drones and support drones to traverse the network given the selected formation and the various constraints. The constraints include different energy consumption rates between the drones due to payload and position in the formation. They also include the different number of recharging stations at every node. 

Algorithm \ref{sequntialAlg} describes the $SDS$ services sequential composition with the in-flight energy sharing process. The swarm computes the shortest path in terms of distance only between the source and destination \cite{dijkstra1959note}, at the source node. If \emph{all} the drones $D = D_d + D_s$ in the swarm $S$ are able to reach the destination without the need to recharge at intermediate nodes, then the swarm will select this path and update the travel time $tt$ (lines 5-7). This capability is computed by considering the amount of consumed energy $E_c$ due to payload, formations, distance, and wind conditions. Otherwise, the swarm estimates the effect of in-flight energy sharing on the swarm to reach the destination directly (line 8). The in-flight energy sharing process consists of two main steps. Namely, \emph{composing EaaS services} and \emph{re-ordering} the drones in the formation when required to facilitate the energy sharing process. These two steps are described in sections \ref{Eaas} and \ref{reordering} respectively.

\begin{algorithm}[!t]
 \caption{Swarm-based Drone Services Composition with In-flight Energy Sharing Algorithm}
 \label{sequntialAlg}
 \small
 \begin{algorithmic}[1]
 \renewcommand{\algorithmicrequire}{\textbf{Input:}}
 \renewcommand{\algorithmicensure}{\textbf{Output:}}
 \REQUIRE $S$, $R$ \hfill\COMMENT{Swarm $S$ and SDS Request $R$}
 \ENSURE  $dt$ \hfill\COMMENT{Total delivery time from source to destination}
 \STATE $dt$ = 0
    \WHILE{$S$ is not at destination}
        \STATE distance to destination= \textbf{Dijkstra}(current, destination)
        \STATE \textbf{compute} $E_c$ for every $D_d$ and $D_s$ in $S$ based on $R$ package weights, $S$ formation, wind condition, and distance to destination
        \IF{all $D$ in $S$ can reach destination without intermediate nodes}
        \STATE $S$ travels to destination
        \STATE $dt$+=$tt$
        \ELSIF{all $D$ in $S$ can reach destination without intermediate nodes using in-flight energy sharing}
        \STATE \textbf{In-flight\_Energy\_Sharing\_Composition()}
        \STATE $dt$+=$tt$
        \ELSE
        \STATE \textbf{find} nearest neighbor nodes %where $lookahead =l$
        \STATE \textbf{EaaS\_Composition()}
        \STATE \textbf{select} best neighboring node, min($tt + ct$)
        \STATE $S$ travels to neighboring node
        \STATE $dt$+=$tt + ct + wt$ 
        \ENDIF
    \ENDWHILE
 \RETURN $dt$
 \end{algorithmic}
 \end{algorithm}

% If the swarm still can't reach the destination even with energy sharing, the swarm needs to stop at intermediate recharging nodes. While the destination is not reachable, the path is composed \emph{sequentially} \cite{alkouz2020swarm}. The swarm selects the nearest reachable node with the least travel time $tt$ and node time $nt$, with in-flight energy sharing in consideration, from the current node. The node time constitutes of the charging time and the waiting times of the drones to charge sequentially $nt=ct+wt$. The charging time $ct$ is the time that the drones take to charge to 100\%. As the energy consumption is different among the swarm, we take the largest charging time $ct$ to be the $wt$ of parallel recharging drones. The node time $nt$ is highly affected by the in-flight energy sharing process before it reaches the node. This is because the drones will reach the node with higher battery capacity as it would be without the energy sharing process. This will save time as it may reduce the charging and waiting time of the drones.  Once the swarm is fully recharged, it tries to find if the destination is directly reachable and repeats the aforementioned steps until it reaches the destination. The total delivery time $dt$ composes of the total travel time $tt$ and the total node times $nt$. Algorithm \ref{sequntialAlg} describes the SDaaS sequential composition with the in-flight energy sharing process.

%=======================================================================================================================

The proposed nested composition comes into play if the swarm cannot reach the destination directly, i.e. without stops at intermediate nodes, even with energy sharing (line 11). The path is composed \emph{sequentially} based on our previously proposed approach \cite{alkouz2020swarm} along with in-flight energy sharing at each line segment in the composed path. The swarm selects the nearest reachable node with the least travel time $tt$ and node time $nt$ (lines 12-14). The node time consists of the charging time and the drone's waiting times to charge sequentially $nt=ct+wt$. 

The charging time $ct$ is the required time to charge the drone to 100\%. As the energy consumption is different among the swarm members, we take the longest charging time $ct$ to be the $wt$ of parallel recharging drones. \textcolor{black}{The waiting time $wt$ highly depends on the number of available recharging pads. When the number of pads is less than the number of drones, an optimization problem takes place to find the best set of drones to charge together to minimize the node time $nt$. Since the number of pads is finite, we use a brute force approach to compute the node time $nt$. For example, if there are 5 drones with charging times $ct$ $<60, 50, 40, 30, 20>$ minutes. Then the node time will be computed as follows:
\begin{itemize}
    \item 1 pad available: node time = 60 + 50 + 40 + 30 + 20 = 200 minutes.
    \item 3 pads available: pad 1: 60 = 60, pad 2: 50 + 20 = 70, pad 3: 40 + 30 = 70, node time = max (60, 70, 70) = 70 minutes.
    \item 5 pads or more available: node time = max (60, 50, 40, 30, 20) = 60 minutes.
\end{itemize}}

The node time $nt$ is highly affected by the in-flight energy sharing process before it reaches the node. The drones will reach the node with higher battery capacity as it would be without the energy sharing process. Once the swarm is fully recharged, it tries to find if the destination is directly reachable and repeats the steps mentioned above until it reaches the destination. The total delivery time $dt$ comprises the total travel time $tt$ and the total node times $nt$ (line 16). 
\subsubsection{In-Flight Energy Sharing Composition}
\label{Eaas}
%Amani: We need to talk about the fact that ERs will be divided in the case of multiple SD 
\begin{figure}[!t]
\centering
  \centering
  \includegraphics[width=\linewidth]{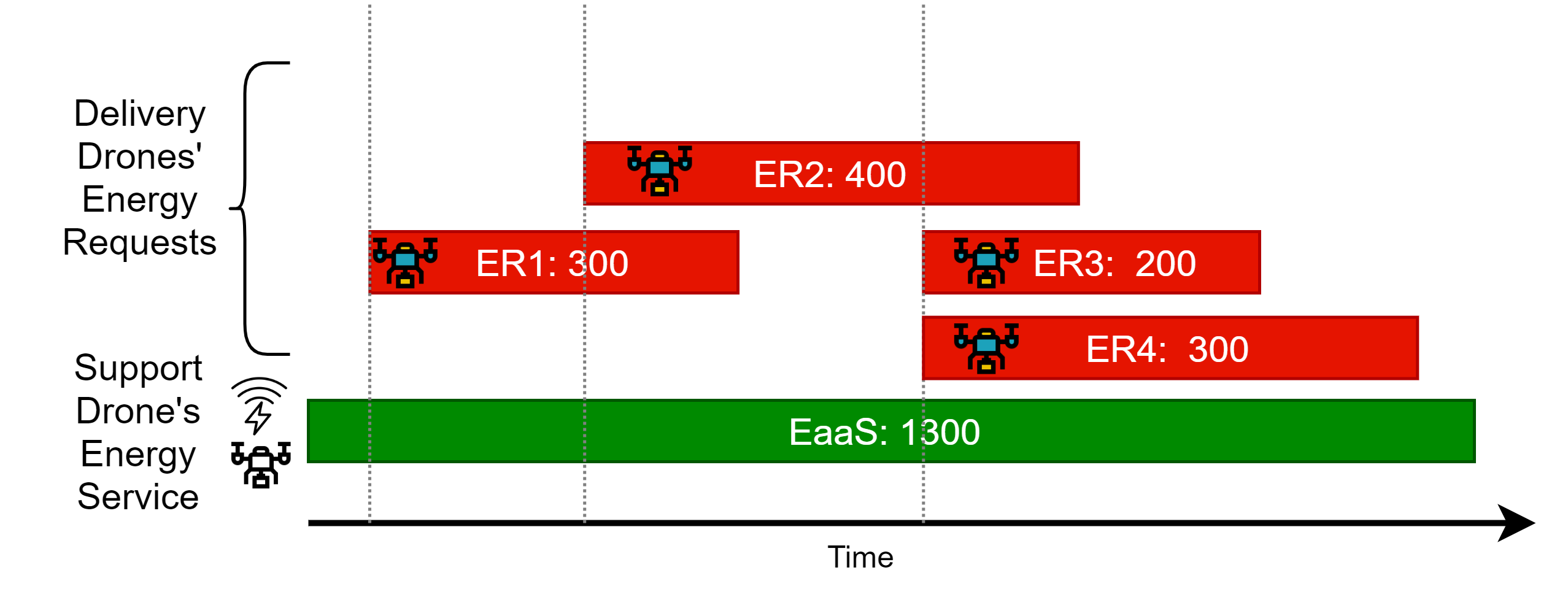}
  \caption{Drones' Energy Sharing Requests and Service }
  \label{fig:ERFull}
\end{figure}

% The in-flight energy sharing composition will be invoked in the path composition as aforementioned. The support drones will share their spare energy with the neighbouring drones in the swarm. In the case of multiple support drones, we assume that each support drone will be responsible to charge a subset of the delivery drones. We also assume that a support drone can share its energy to one delivery drone at a time \cite{lakhdari2018crowdsourcing}\cite{lakhdari2020Elastic}. The energy sharing time window is the travel time between the current node and the next destination node. An energy sharing request will be triggered by the delivery drones. As mentioned earlier, each drone has its own energy consumption model. Once the battery capacity of a drone becomes lower than a threshold $\alpha$, the in-flight energy sharing is requested. The in-flight energy sharing composition aims to select and compose the energy requests that will maximize the utilization of support drones' spare energy. Maximizing the energy utilization will reduce the required time to wait and charge from charging pads in an intermediate node. In this paper, we propose to compose energy requests using the below two approaches.

% Both approaches are inspired by the first come first served scheduling algorithm \cite{kruse2007data}. Thus the priority in selecting energy requests will be impacted by their starting time $st$.

%=========================================================================================================================================

The in-flight energy sharing composition will be invoked in the path composition. The support drones will share energy with the neighboring drones in the swarm. In the case of multiple support drones, we assume that each support drone will be responsible for charging a subset of the delivery drones. We also assume that a support drone can share its energy to one delivery drone at a time \cite{lakhdari2020composing}\cite{lakhdari2020Elastic}. The energy sharing time window is the travel time between the current node and the next destination node, i.e., a line segment in the composed $SDS$ services path. The delivery drones will trigger an energy sharing request. \textcolor{black}{Each drone has a different energy consumption model depending on several factors. The factors include the distance travelled, carried payload, wind condition, and the drones position within a formation. If any of these factors change during the composition, the energy consumption for each drone is recomputed.} Once the drone's battery capacity decreases to a predefined threshold $\gamma$, an in-flight energy sharing request is launched. The in-flight energy sharing composition aims to select and compose the energy requests $ER$ that will maximize the utilization of support drones' energy. Maximizing the energy utilization will reduce the required time to wait and charge at charging pads in an intermediate node. In this paper, we propose to compose $ERs$ using two approaches and study their impact on the travel time. The first approach, i.e. Priority-based (PB) energy sharing, is inspired by the first come first served scheduling algorithm \cite{kruse2007data}.  In PB approach, the support drone charges drones based on the start time of their request. Moreover, the support drone fully charge each selected drone based on their requests. The second approach, i.e. Fairness-based (FB) energy sharing, charges each drone partially in a round robin manner. %We are studying the impact of each  approach on the travel time. 

%\begin{itemize}%[leftmargin=10pt,labelsep=5pt,itemindent=5pt, labelwidth=*]
    %\item 
    
    \begin{algorithm}[!t]
 \caption{In-flight PB Energy Sharing Composition }
 \label{FullEaaSAlg}
 \small
 \begin{algorithmic}[1]
 \renewcommand{\algorithmicrequire}{\textbf{Input:}}
 \renewcommand{\algorithmicensure}{\textbf{Output:}}
 \REQUIRE $battery[D], tt$ %\hfill\COMMENT{$battery[D]$: delivery drones' battery, $tt$: segment travel time}
 \ENSURE  $updated\_battery[D]$
 \STATE $ER$ = \textbf{Generate\_ER($battery[D]$)}
 \STATE $SortedER$ = \textbf{sort}($ER$, $st:ascending$, $ re:descending) $
 \STATE $st = tt.st$
 \STATE $et = tt.et$ 
 \FOR{$er_{i} \in SortedER$}
    \IF{$er_i.st\geq SD.st$ \AND $er_i.re\leq SD.ae$}
        \STATE $st = er_i.et$
        \STATE $SD.ae = SD.ae - er_i.re $
        \STATE $ER\_composition$.\textbf{add}($er_i$)  
        \STATE $battery[D] = battery[D] + er_i.re $
        \STATE $SortedER$.remove($er_i$)
        \STATE $battery[D]  = $ \textbf { Reoder(Fixed/Flexible) } 
        \STATE $newER$ = \textbf{Generate\_newER}($battery[D]$)
        \IF{$newER \neq \emptyset$}
            \STATE $SortedER = SortedER \; \cup \; newER$
            \STATE $SortedER$ = \textbf{sort}($SortedER, st:ascending,$ $ re:ascending) $
        \ENDIF
    \ENDIF
 \ENDFOR
 \STATE $updated\_battery[D] = battery[D]$
 \RETURN $updated\_battery[D]$
 \end{algorithmic}
 \end{algorithm}
\subsubsection*{Priority-Based (PB) Energy Sharing}
 This approach is a modified version of the first come first served scheduling algorithm \cite{kruse2007data}. Thus, the priority in selecting $ERs$ will be defined by the requests' starting times $st$. In this approach, selected $ERs$ will be charged fully according to their requested amount. Each delivery drone requesting energy defines the starting time ${st}$ and the requested amount ${re}$ for their request. If the support drone has enough spare energy to provide and multiple requests overlaps, the earlier request will be selected. If multiple requests start simultaneously, then the request with the highest requested energy will be selected. For instance, using this approach on the given requests in Fig. \ref{fig:ERFull}, ER1 is selected since it is the earliest request. %ER2 will be skipped because the support drone is still charging the drone of ER1. Lastly, ER4 is selected because it requests more energy than ER3 even though both requests start at the same time. 
    
  Algorithm \ref{FullEaaSAlg} presents the PB energy sharing composition in detail. The algorithm takes as an input the delivery drones' battery $battery[D]$ and the segment travel time $tt$. The algorithm returns as an output the updated delivery drones' battery $updated\_battery[D]$. Line 1 retrieves the available energy requests ($ERs$) by checking the battery of the delivery drones. If the drone's battery capacity decreases during the flight on the line segment then an energy request ($er$) is launched. The $er$'s start time will be when the drone's battery reaches less than a predefined threshold $\gamma$. Line 2 sorts the set of $ERs$ in ascending order based on their start time, then in descending order based on the requests' size. In Lines 3 - 19, For every $er$, the algorithm checks if: (1) the start time of the $er$ falls in the time period of the support drone service (2) the support drone has enough energy to fulfill the required energy, then the $er$ will be selected. Once an $er$ is selected, the support drone time and capacity will be updated (Lines 7-8). In addition, the $er$ information will be added to the composition (Line 9). Line 10 updates the drone's battery of the selected $er$. Line 11 removes the $er$  from the set of $ERs$.  Moreover, a formation re-order function will be called (Line 12). The re-order function checks if a re-positioning of the drones is required to deliver the requested energy. The position of the drones affects their energy consumption which may trigger new $ERs$ that should be considered in the current in-flight energy sharing composition. Therefore, re-ordering our energy sharing composition approach will check if new $ERs$ are submitted by the delivery drones (Lines 13 - 17). The formation re-ordering approach is discussed in detail in section \ref{reordering}. \looseness=-1
    
    %\item \textbf{Fixed Energy Sharing: }\hl{Do we really need this one?}

% Amani: Do we need a new figure?
% \begin{figure}[!t]
% \centering
%   \centering
%   \includegraphics[width=\linewidth]{images/DronesERPartial.png}
%   \caption{Drones' energy requests for In-flight Partial Energy Sharing }
%   \label{fig:ERRR}
% \end{figure} 

    %  \item \textbf{Round Robin Energy Sharing: }In this approach, selected energy requests may be charged partially based on their start time $D_{st}$, the available energy requests $ER$, and their requested amount of energy $D_{re}$.  Requests will be chunked to maximize the utilization of the spare energy of the support drone. The chunks are defined based on the available energy requests start and end times \cite{lakhdari2020composing}. Then, chunks will be selected based on starting time $c_{st}$ and requested amount of energy $c_{re}$. If multiple chunks overlap in the same time frame, then the chunk with the highest requested energy will be selected. For example, using this approach on the same given requests in Fig. \ref{fig:ERFull} will result in the chunks represented in Fig.\ref{fig:ERPartial}. Additionally, the set of selected and composed chunks using this approach will be \{ER1C1, ER2, ER3C2, ER4C3\}. Notice that ER1C1 (ER1 chunk1) is selected as it's the only available chunk of a request. ER2C1 is selected because it requests more energy than ER1C1. The rest of ER2 is selected because ER2 chunks request more energy than any other overlapping chunks. ER3C2 and ER4C3 are selected for the same reason as ER2. 
     
     \subsubsection*{Fairness-Based (FB) Energy Sharing} In this approach, a fixed amount of energy will be provided to each drone in a round robin fashion based on the drone's id until the end time of the segment.  
      The amount of energy may vary from small to large amount. %We test this approach with different sizes of provided energy to study its impact on the SDaaS composition. 
      Algorithm \ref{RREaaSAlg} describes the Fairness-based in-flight energy sharing composition. The amount of provided energy is fixed based on a user-defined value $\lambda$ (Line 3). Lines 4, computes the charging time based on the charging rate of the support drone. The algorithm then goes through drones in order of ID to provide a fixed amount of energy (Lines 5 -16). Note that the energy will be provided by the support drone in rotation between the delivery drones as long as: (1) the support drone battery is greater than a threshold $\delta$. the threshold \textcolor{black}{$\delta$ is} used to keep enough spare energy for the support drone's own energy consumption. (2) there is enough time to charge a delivery drone. Similar to the Priority-Based energy sharing approach, once a drone is selected (Lines 5 - 11), the battery of all drones will be updated (Lines 9 - 11) and a formation re-order function will be called (Line 12). The re-order function checks if a re-positioning of the drones is required to deliver the requested energy. As previously mentioned, the positions of the drones affect their energy consumption which needs to be updated in the current in-flight energy sharing composition. The formation re-ordering approach is discussed in detail in section \ref{reordering}.

\begin{algorithm}[!t]
 \caption{In-flight FB Energy Sharing Composition}
 \label{RREaaSAlg}
 \small
 \begin{algorithmic}[1]
 \renewcommand{\algorithmicrequire}{\textbf{Input:}}
 \renewcommand{\algorithmicensure}{\textbf{Output:}}
 \REQUIRE $battery[D], tt$
 \ENSURE  $updated\_battery[D]$
 %\STATE $ER$ =\textbf{ Generat\_ER}($battery[D]$)
 %\STATE $SortedER$ = \textbf{sort}($ER, st:ascending,$ $ re:descending) $
 \STATE $ct = tt.st$
 \STATE $et = tt.et$  
 \STATE $ amountE = \lambda$
 \STATE $chargingt = amountE/chargingRate$

\WHILE{$ct < et$ \AND $ SD.ae > \delta$}
    \FOR{$d_{i} \in battery[D]$}
    \IF{$ct < et$ \AND $ SD.ae > \delta$}
        \STATE $ct = ct + chargingt$
        \STATE $SD.ae = SD.ae - amountE$
        \IF{$battery[D]$ \textbf{is not full}}
            \STATE $battery[D] = battery[D] + amountE$
            \STATE $battery[D]  = $ \textbf { Reorder(Fixed/Flexible) } 
        \ENDIF
    \ENDIF
    \ENDFOR
\ENDWHILE
\STATE $updated\_battery[D] = battery[D]$
 \RETURN $updated\_battery [D]$
 \end{algorithmic}
 \end{algorithm}

\subsubsection{Formation Re-ordering} 
\label{reordering}
Once the in-flight energy sharing process gets initiated, the swarm may need to re-order itself to facilitate the process. The support drones, as described earlier, could share their energy to up to 1.2 meters \cite{benbuk2020leveraging}. Hence, the support drones should neighbor the energy requesting delivery drone. Therefore, the drones in the swarm swap positions to facilitate the process. For simplicity, we assume that there is no overhead time and energy during re-ordering. However, the energy consumption rate of the drones will change due to their position change in the formation. \textcolor{black}{We recompute the new energy consumption model whenever re-ordering occurs.} In this paper, we investigate a fixed approach for swarm re-ordering. 

%The support drones are assumed to be in fixed positions.

In a fixed approach, the support drones do not change their positions. Once the energy sharing is initiated and the distance between the support drone and the energy requesting delivery drone is more than 1.2m, the delivery drones swap their places to neighbor the support drones. Otherwise, the support drone directly shares its energy without re-ordering. In this method, we ensure that the energy consumed by the support drone does not change. If the support drone is placed in the optimal position initially, it will be consuming the least amount of energy due to its position. On the other hand, the delivery drones swapped may be placed in a non-optimal position during the energy sharing process. Once the energy sharing process ends, the drones re-order themselves again to their initial positions.

\section{Experiments and Results}
\label{expirementsandresults}

We evaluate the proposed swarm-based drone services composition framework with in-flight energy sharing.  We assess the proposed framework against three composition algorithms, baseline \cite{alkouz2020swarm}, Dijkstra's \cite{dijkstra1959note}, and \textcolor{black}{Floyd-Warshall's \cite{sung2020multi}.} We first conduct a set of experiments to evaluate the \emph{effectiveness} and \emph{feasibility} of the system under the different controlling attributes, including the initial positioning and energy sharing methods. We assume the amount of energy shared per round for the Fairness-based energy sharing method is  2240mA. We first compare the number of fulfilled requests under the different energy sharing methods. Second, we compare the delivery times of the proposed requests under different settings. Finally, we compare the runtime efficiency of the composition methods.

\subsection{Dataset and Experiments Setup}
% \usepackage{graphicx}
\begin{table}[]
\caption{Experimental Variables}
\label{tab:variables}
\resizebox{\linewidth}{!}{%
\begin{tabular}{l|l}
\hline
Variable                                      & Value         \\ \hline
No. of nodes in the network subset            & 276           \\
No. of nodes in the largest connected network & 195           \\
No. of generated requests                    & 10000          \\
Max weight of a package                       & 1.4 Kg        \\
Drone model                                   & DJI Phantom 3 \\
Speed of the drone                            & 30 km/h      \\
Battery capacity                              & 4480 mAh       \\
Voltage                                       & 15.2 V        \\
Rate of in-flight energy sharing                        & 5.88 mAh/min    \\ \hline
\end{tabular}%
}
\end{table}
To the best of our knowledge, there is no publicly available data of drone trajectories in skyway networks. Therefore, the dataset used for the experiments is an urban road network dataset from London city to mimic a skyway arrangement. The nodes are the intersections in the city, and the edges represent the distances between the nodes \cite{karduni2016protocol}. We took a sub-network of connected nodes with the size of 195 nodes to mimic how a skyway network may look like for our experiments. We then synthesize 10000 requests with random source and destination nodes. We also synthesize different wind speeds and directions for the skyway segments. We only consider wind speeds under 13.8m/s that drones are safe to fly in \cite{alkouz2020formation}. \textcolor{black}{We have incorporated a real drone trajectory dataset \cite{jermaine2021demo} used in the energy model calculations. The dataset however has limitations of being only for single drones. Hence, it does not reflect the formations effect on the energy consumption due to drag and upwash/downwash forces. Therefore, we used the payload and flight range effect on the energy consumption from the dataset. We then augmented the real dataset with the computational fluid dynamics study proposed in \cite{alkouz2020formation} to learn the behaviour in terms of swarm formation and the wind effect on the energy consumption. The maximum weight of a package is assumed to be 1.4 kg following the payload capacity of DJI Phantom 3 used in the experiments. The real-drone dataset uses the CrazyFlie 2.1 \footnote{https://www.bitcraze.io/products/old-products/crazyflie-2-0/} drone which is a small drone with 240mAh and 3.7 V battery compared to the DJI Phantom 3 with 4480mAh and 15.2 V battery. Therefore we multiply the consumption rate by the increasing factor, i.e. 4.1.}  Table\ref{tab:variables} summarizes the experimental variables. The variables are used to compute the energy consumption.

%The upwash/downwash and drag forces are generated for the same drone model using Ansys Discovery Live\footnote{Ansys Discovery Live. https://www.ansys.com/products/3d-design/ansys-discovery-live} and reported in \cite{alkouz2020formation}.  

\textcolor{black}{We compare the proposed composition algorithm against a baseline approach \cite{alkouz2020swarm}, Dijkstra's algorithm \cite{dijkstra1959note} and Floyd-Warshalls algorithm \cite{sung2020multi}.} The baseline approach considers all aspects of composition without the in-flight energy sharing. Formations, intrinsic constraints,  and extrinsic constraints are all considered in the baseline approach. Hence, we evaluate across the two levels. The blue parts of Fig. \ref{fig:Composition_framework} represent the baseline approach. \textcolor{black}{For Dijkstra's and Floyd-Warshall's methods, we consider the cost of every edge to be the travel time $tt$ of the edge and the node time $nt$ (charging + waiting times).} The proposed framework is evaluated under the two different positioning settings of the support drones. The first setting positions the support drones at the worst positions (location-aware). The second setting positions the support drones in the best positions (energy-aware). In addition, the proposed framework is evaluated using the different energy sharing methods proposed, i.e., Priority-based (PB) and Fairness-based (FB) sharing.

% In the fourth experiment, we compare the proposed composition algorithm against a baseline approach and Dijkstra's algorithm \cite{dijkstra1959note}. The baseline approach considers all aspects of composition without the in-flight energy sharing. Formations, intrinsic constraints,  and extrinsic constraints are all considered in the baseline approach. The blue parts of Fig.\ref{fig:Composition_framework} represent the baseline approach. For Dijksta's method we consider the cost of every edge to be the travel time $tt$ of the edge and the node time $nt$ (charging + waiting times). As shown in Fig. \ref{fig:successful}, our proposed energy sharing approach performed much better than the baseline and Dijkstra's. In baseline, only 8839 requests are successful out of 10,000 which is less than any of the 4 combinations explained in experiment 1. Dijkstra's performed worse in terms of successful requests (6,646 successful out of 10,000). This is because our modified A* sequential composition algorithm can detour its path if the swarm would get stuck at a node. In contrast, Dijkstra's composition of the path is solely based on the cost of each segment only. Hence, the composed path may have segments that a swarm can't deliver due to its battery limitations. In terms of delivery times, the in-flight energy sharing compositions outperform the baseline under best positioning approach (Fig. \ref{fig:App2dt}). This reflects the usefulness of implementing in-flight energy sharing to overcome a swarms battery limitation and increase its flight range.
% \vspace{-8pt}
\subsection{Effectiveness}
\begin{figure}[!t]
\centering
\setlength{\abovecaptionskip}{1pt}
  \setlength{\belowcaptionskip}{-15pt}
  \includegraphics[width=0.95\linewidth]{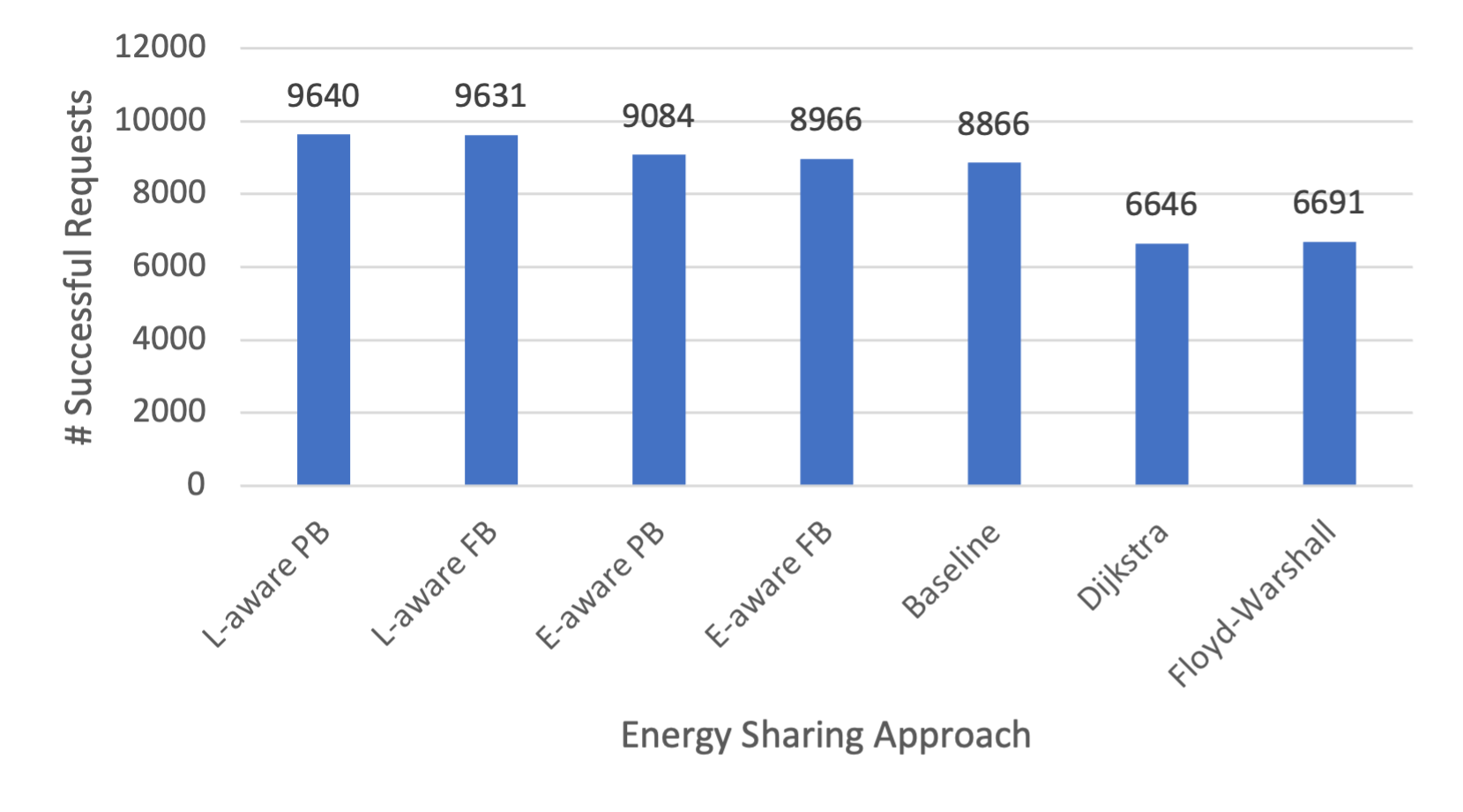}
  \caption{\textcolor{black}{Number of Successful Requests Using In-Flight Energy Sharing}}
  \label{fig:successful}
\end{figure}

In the first experiment, we compare the number of successful requests between the different settings proposed. A successful request is a request that was able to be served by the swarm without getting stuck in intermediate nodes due to lack of power. \textcolor{black}{We compare the 4 different combination of settings against the Baseline, Dijkstra's, and Floyd-Warshall's algorithms.} This includes the initial positioning, and energy sharing approaches. As shown in Fig. \ref{fig:successful} , the trend shows that \emph{positioning} the support drones in the optimal position (energy-aware) generally results in less number of successful requests \textcolor{black}{(18,050 successful requests out of 20,000 compared to 19,271 in location-aware setting).} This is because positioning the support drones in the optimal position means the rest of delivery drones will be shifted to worst positions. The number of delivery drones is usually higher than the support drones and hence more energy consumption occurs during the travel. Although in this setting the support drones consume the least amount of energy and is able to share more of its excess energy with the delivery drones, the number of energy requests generated by the delivery drones is more than what the support drones can serve at a time. Hence, more failures occur. Last, we can note that the PB energy sharing approach performs better in all settings. All requests are successfully delivered compared to the Fairness-based energy sharing. This finding is explained in the second experiment. \textcolor{black}{In comparison with the baseline, Dijkstra's, and Floyd-Warshall, our proposed energy sharing approach performed much better. In baseline, only 8839 requests are successful out of 10,000 which is less than any of the 4 combinations. This is because a swarm in the baseline approach solely depends on the delivery drones batteries and in many cases are stuck at a node due to the battery limitations. Floyd-Warshall's performed worse than the baseline but slightly better than Dijkstra's (6,691 successful out of 10,000). Dijkstra's performed worse in terms of successful requests (6,646 successful out of 10,000).} This is because our modified A* sequential composition algorithm can detour its path, in the baseline and in the energy sharing methods, if the swarm would get stuck at a node. \textcolor{black}{In contrast, Dijkstra's and Floyd-Warshall's composition of the path is solely based on the cost of each segment only.} Hence, the composed path may have segments that a swarm can't travel in due to its battery limitations. \looseness=-1

% We can also note that a Fixed \emph{re-ordering} in general results in more successful requests compared to the Flexible re-ordering. This is specially significant in Approach 1 with Partial energy sharing where the Fixed re-ordering resulted in 9999 successful requests and the Flexible re-ordering resulted in 8869 successful requests only. This proves that support drones are best placed at their initial position and the delivery drones move when needed. This approach stabilizes and maintains the support drone battery status. 

\begin{figure*}[ht]
\centering

\subfloat[Support Drones in Worst Positions (Location-Aware)]{\includegraphics[width=0.4\linewidth]{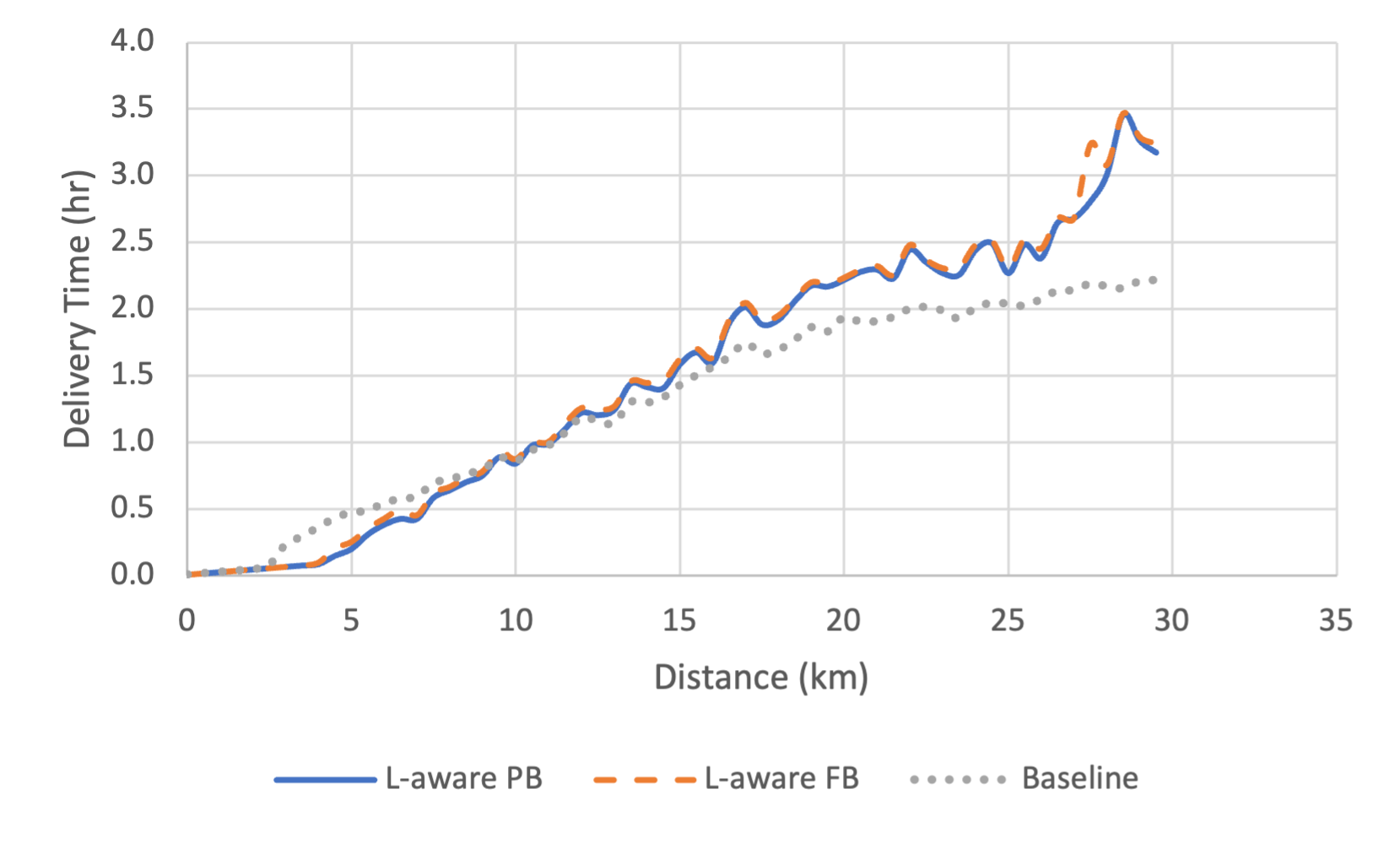}\label{fig_app1dt}\setlength{\belowcaptionskip}{-5pt}
}\quad
\subfloat [Support Drones in Best Positions (Energy-Aware)]{\includegraphics[width=0.4\linewidth]{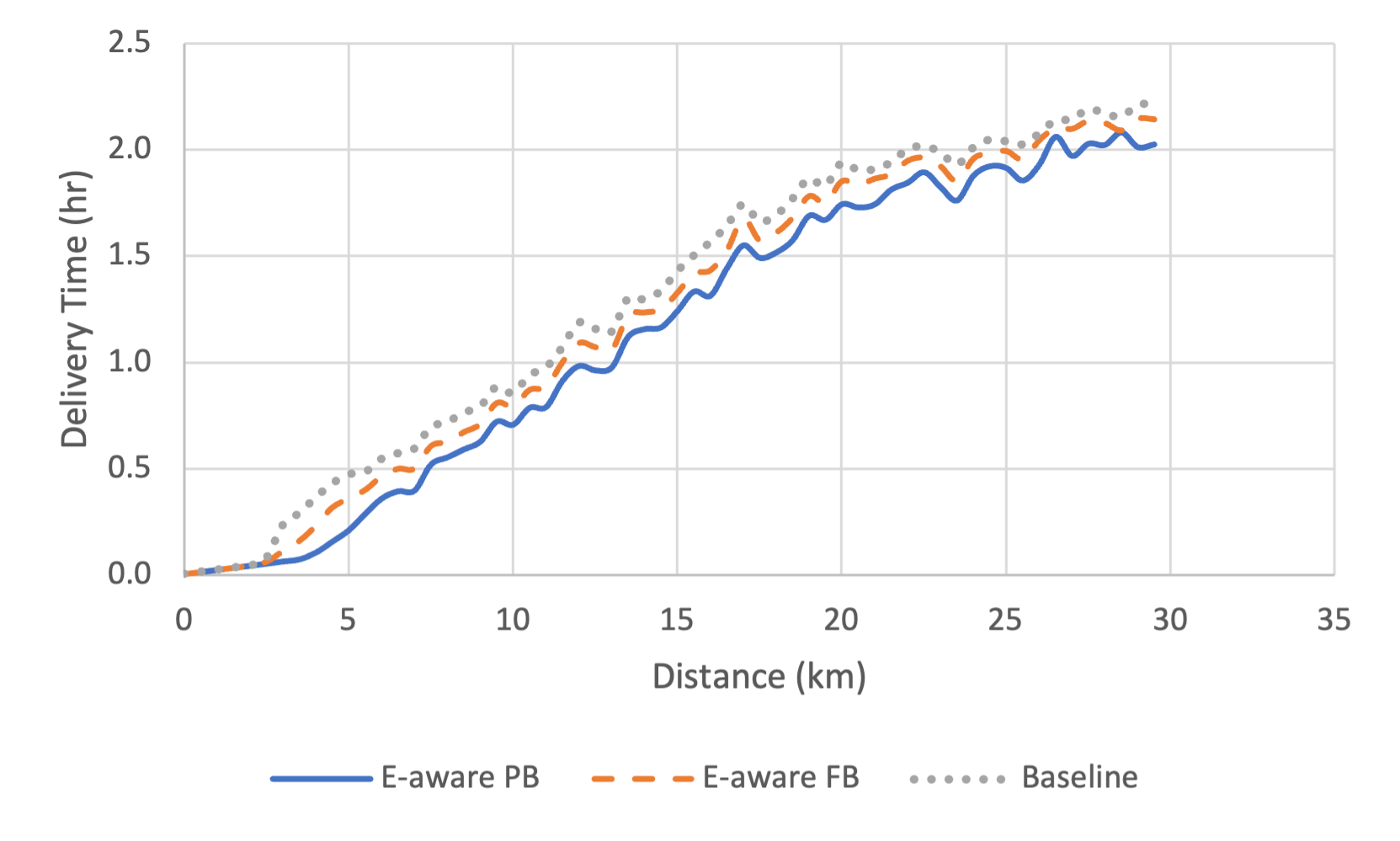}\label{fig_app2dt}\setlength{\belowcaptionskip}{-5pt}
}
\setlength{\belowcaptionskip}{-10pt}
\caption{\textcolor{black}{Delivery Times of Various Energy Sharing Methods}}
\label{fig:Delivery}
\vspace{-3mm}
\end{figure*}

\begin{figure*}[ht]
\centering

\subfloat[Delivery Times of PB Energy Sharing]{\includegraphics[width=0.4\linewidth]{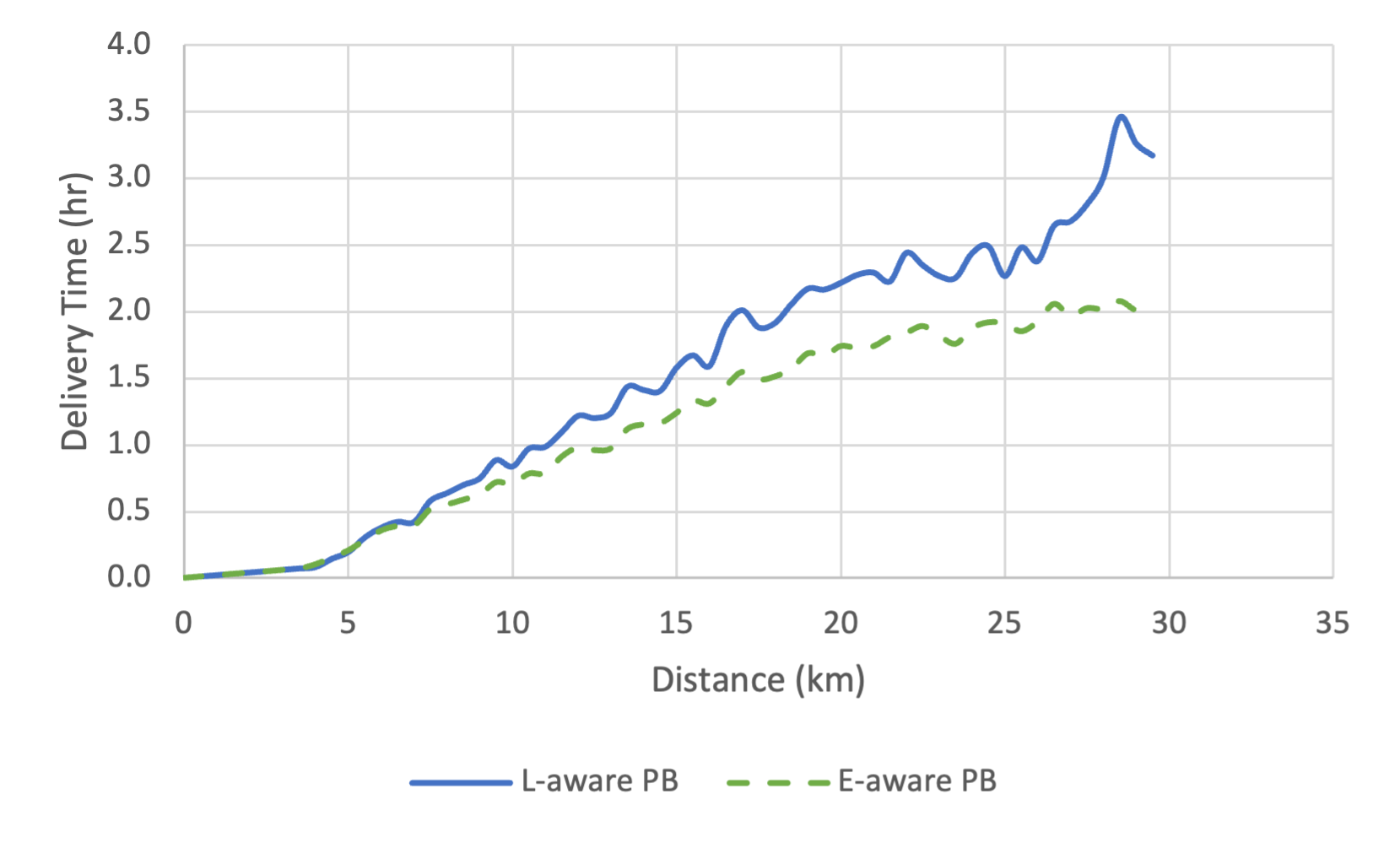}\label{fig_full}\setlength{\belowcaptionskip}{-5pt}
}\quad
\subfloat[Delivery Times of Fairness-Based Energy Sharing]{\includegraphics[width=0.4\linewidth]{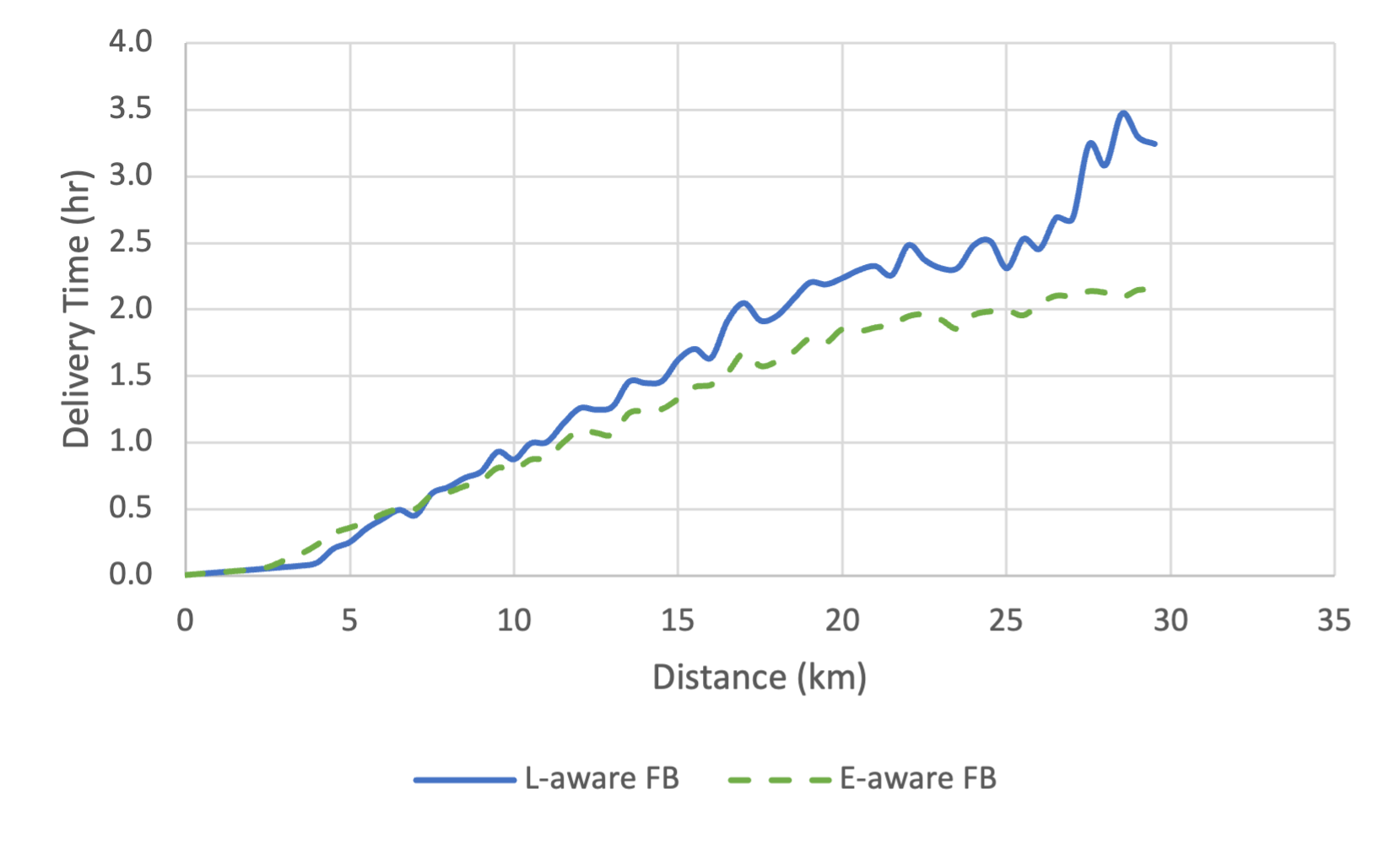}\label{fig_rr}\setlength{\belowcaptionskip}{-5pt}
}
\setlength{\belowcaptionskip}{-10pt}
\caption{\textcolor{black}{Delivery Times Under Different Positioning Settings}}
\label{fig:approaches}
% \vspace{-3mm}
\end{figure*}

In the second experiment, we focus on studying the behavior of the two proposed energy sharing methods, i.e Priority-based (PB) and Fairness-based (FB). Figs. \ref{fig_app1dt} and \ref{fig_app2dt} present the delivery times of the successful requests in PB and FB sharing under the different positioning settings. The x-axis presents the distance in kilometers and the y-axis is the delivery time in hours. The successful requests are grouped by the distance between the source and the destination at 0.5km intervals. The average delivery time of requests in every interval is presented on the graphs. As shown in the figures, the PB sharing is steadily performing the best in the different settings. This is because the energy sharing requests by the delivery drone are generated dynamically after every served request (refer to Algorithm \ref{FullEaaSAlg}). This ensures proactive dynamic generation of energy requests. Hence, most needy drones are served first which ensures better distribution of energy. Better distribution of energy means less time spent at intermediate nodes to charge and may also mean the ability of skipping nodes if the swarm can reach the destination directly without stopping. Hence, a better delivery time. With FB, the delivery drones are equally served at every $i$ interval. This means that there is no priority in charging. Most needy delivery drones may be served later or might not get the chance to be charged if the skyway segment is short. As explained earlier, the energy sharing process is constrained by the travel time of an edge. The better performance of PB sharing is more visible in the energy-aware setting where the support drones are positioned in the optimal positions and the delivery drones are losing more energy due to their positions. This means that the delivery drones are generating more energy requests in this positioning approach and the difference between the energy sharing methods is more evident.

\begin{figure*}[ht]
\centering

\subfloat[Support Drones in Worst Positions (Location-Aware)]{\includegraphics[width=0.4\linewidth]{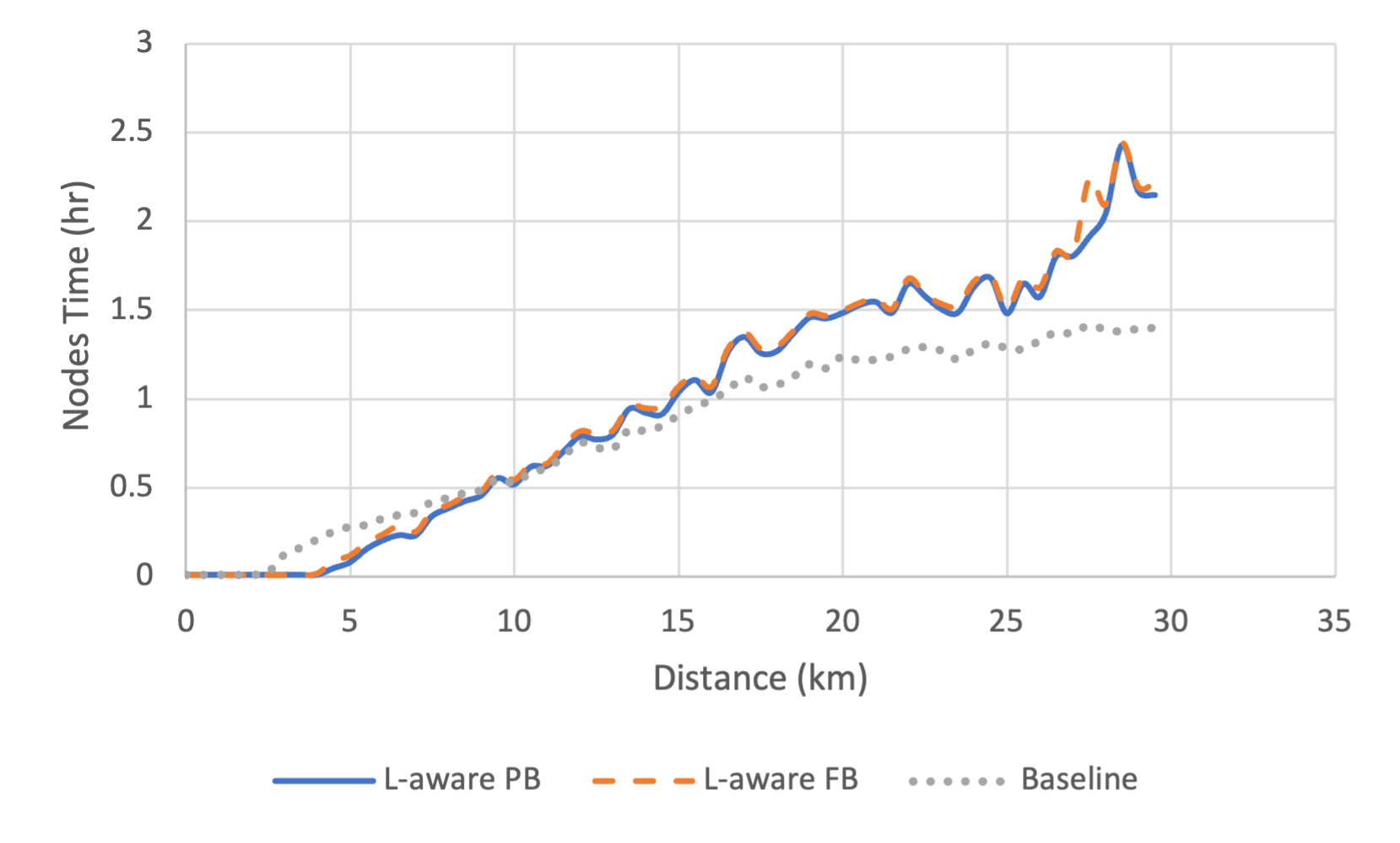}\label{fig_app1nt}\setlength{\belowcaptionskip}{-5pt}
}\quad
\subfloat [Support Drones in Best Positions (Energy-Aware)]{\includegraphics[width=0.4\linewidth]{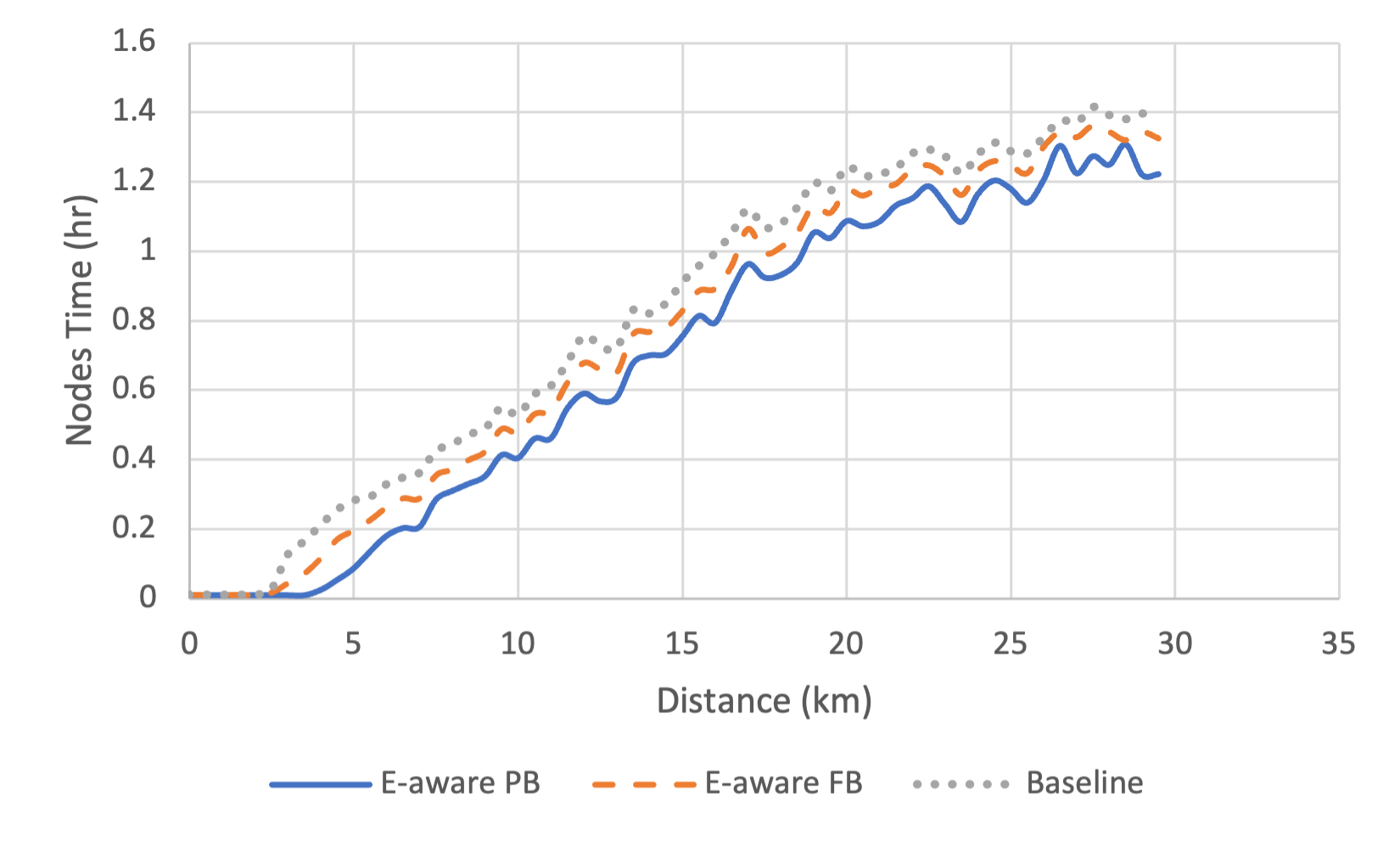}\label{fig_app2nt}\setlength{\belowcaptionskip}{-5pt}
}
\setlength{\belowcaptionskip}{-10pt}
\caption{\textcolor{black}{Node Times of Various Energy Sharing Methods}}
\label{fig:node}
% \vspace{-3mm}
\end{figure*}

In the third experiment, we study the initial positioning settings and their effect on the delivery times. As shown in Fig. \ref{fig_app1dt}, positioning the support drones in the worst position (location-aware) results in worst delivery times compared to the baseline at longer distances although it had the most number of successful requests (Fig. \ref{fig:successful}). When the swarm lands on a recharging station, all the drones charge fully. If the support drones are in the worst positions they consume the most energy (in addition to given energy in the sharing process). Since the support drones have bigger battery capacities than delivery drones, the time taken to recharge fully will be higher when landing as the rest of the swarm is waiting for it. This is especially significant at longer distances as shown in Fig. \ref{fig_app1dt}. \textcolor{black}{This behaviour is reflected in Fig. \ref{fig_app1nt} that represent the nodes time with the location-aware setting. The time spent at the node is higher than the baseline due to the charging of the support drones which are placed in the worst position consuming the most amount of energy.} With longer distances, the support drone consumes more energy and shares more energy, hence, spends more time charging. Whereas in an energy-aware setting, where the support drone is positioned in the optimal position, it takes less time to recharge at a recharging station. The delivery times, are therefore, better than the location-aware setting. This difference is visible in Figs. \ref{fig_full} and \ref{fig_rr} with both PB and FB energy sharing. Hence, we conclude that based on the goal of the application, the choice of the setting in positioning is determined. If the goal is to serve as many requests as possible, the support drones would be placed in the worst positions. Otherwise, if the goal is to serve requests as fast as possible, the support drones would be placed in the best positions. For example, in emergency cases the delivery time would be the most critical objective. If the delivery is successful then the energy-aware setting should be adopted, otherwise the location-aware setting would be adopted to ensure the success of the delivery. Moreover, the in-flight energy sharing composition outperforms the baseline under the best positioning setting (Fig. \ref{fig_app2dt}). This reflects the usefulness of implementing in-flight energy sharing to overcome a swarms' battery limitation and increase its flight range. \textcolor{black}{Figs. \ref{fig_app1nt} and \ref{fig_app2nt} represent the node times of the delivery requests without the travel times. The figures prove that the improvement in delivery times is due to the reduction of node times because of in-flight recharging. The behaviour of the lines are similar to Figs. \ref{fig_app1dt} and \ref{fig_app2dt} but with lower times demonstrating the reduction in node times mainly.}

\subsection{Runtime Efficiency}

\begin{figure*}[ht]
\centering

\subfloat[Support Drones in Worst Positions (Location-Aware)]{\includegraphics[width=0.4\linewidth]{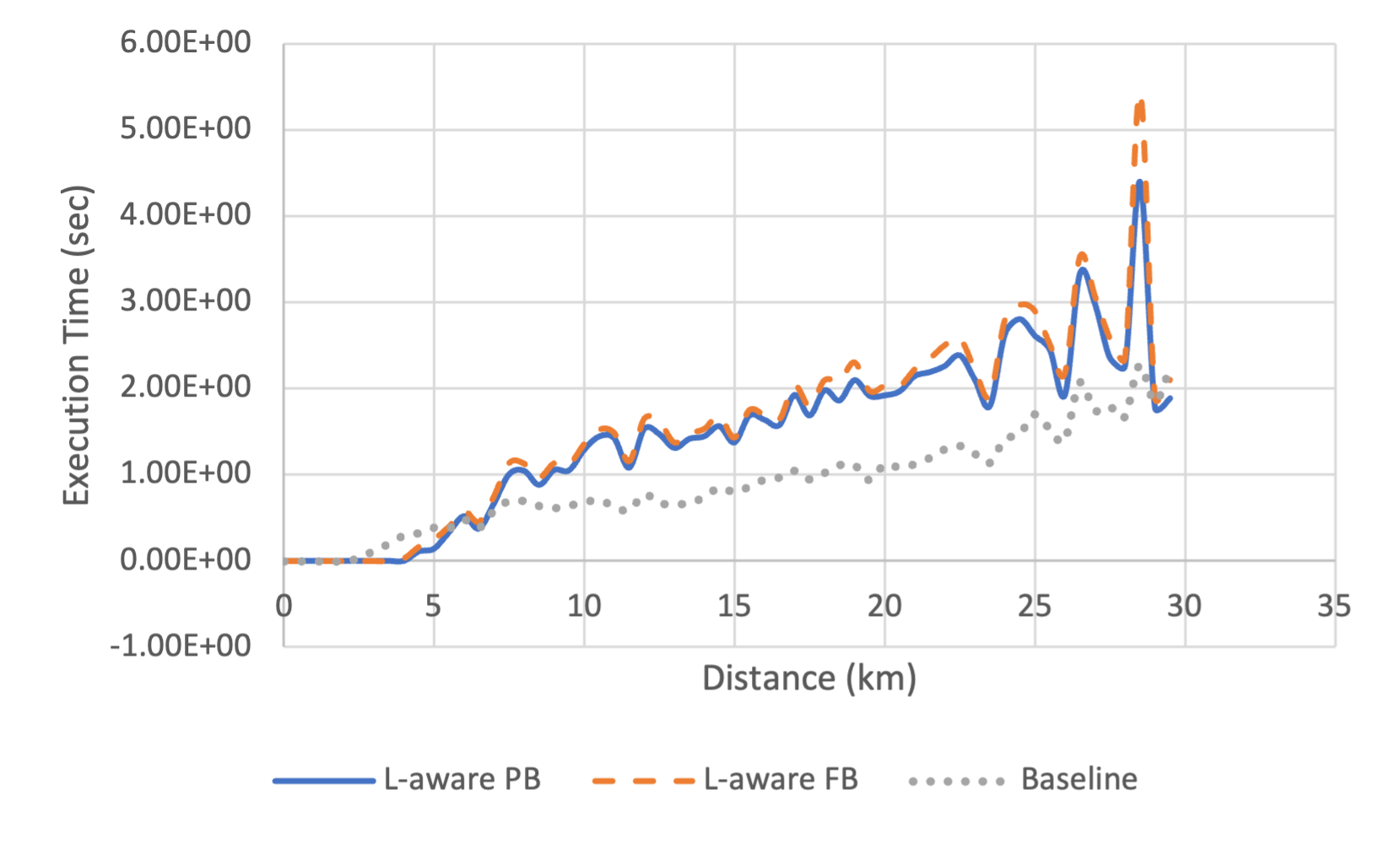}\label{fig_app1et}\setlength{\belowcaptionskip}{-5pt}
}\quad
\subfloat [Support Drones in Best Positions (Energy-Aware)]{\includegraphics[width=0.4\linewidth]{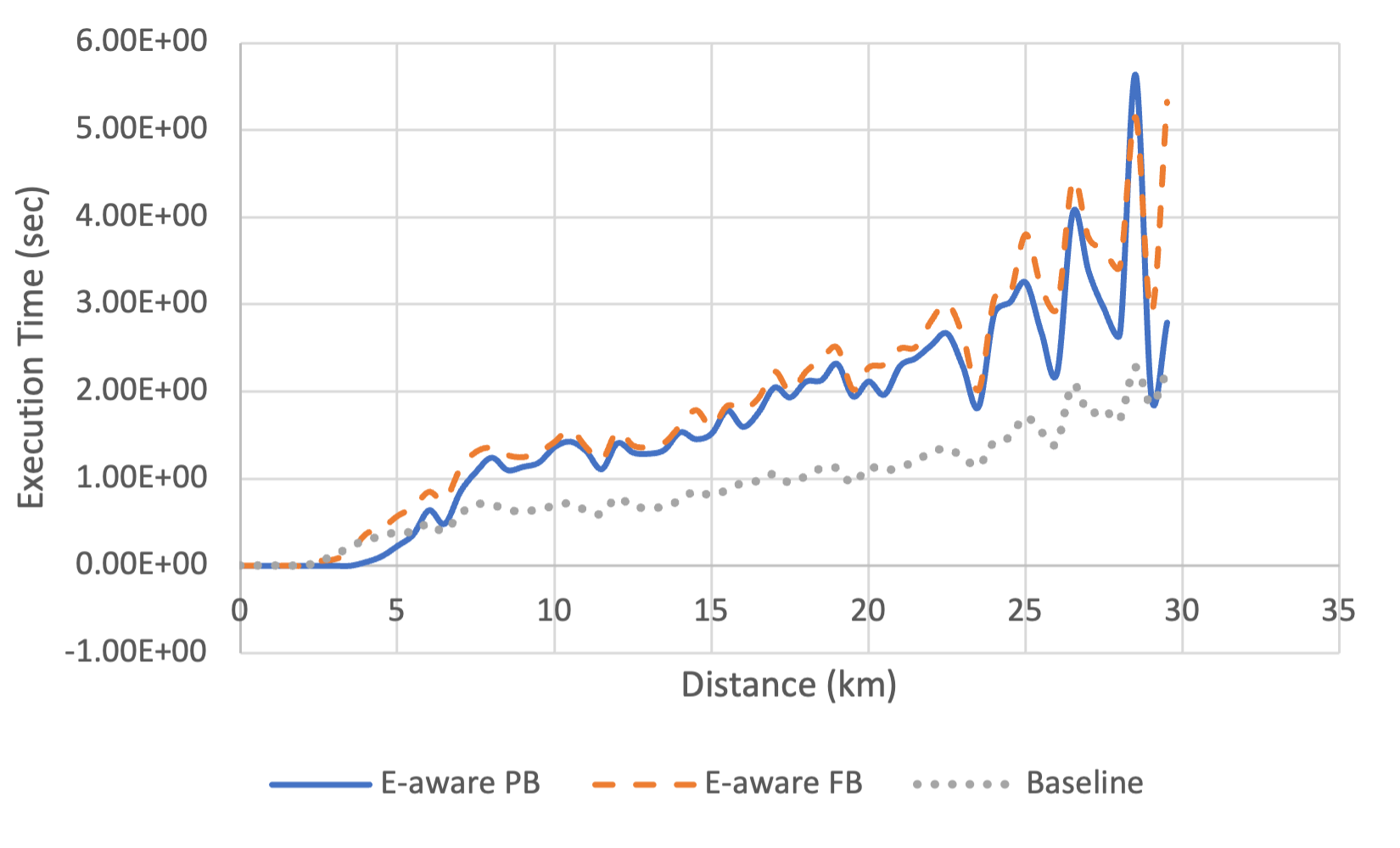}\label{fig_app2et}\setlength{\belowcaptionskip}{-5pt}
}
\setlength{\belowcaptionskip}{-10pt}
\caption{\textcolor{black}{Execution Times of Various Energy Sharing Methods}}
\label{fig:Execution}
\vspace{-3mm}
\end{figure*}

We measure the execution times of the proposed methods. Figs.\ref{fig_app1et} and \ref{fig_app2et} depict the execution times of different positioning and energy sharing approaches. As shown in the graphs, the positioning does not affect the execution times. However, in both positioning settings, the execution time of the PB energy sharing method outperforms the FB energy sharing methods. This is because the PB sharing only gives energy on request. The FB is pre-emptive and keeps charging the drone iteratively by switching between them, increasing the algorithm's execution time. In addition, with the FB energy sharing, the shared amount per round is parameterized. This means the amount of shared energy per round is input by the user. The smaller the amount is, the more rounds there are which would make the execution time longer. For this experiment, we assume the support drones share 2240 mA per round. If the amount is smaller, less time would be needed to share the energy and therefore more rounds in the FB execution. The baseline's execution time is less than any proposed method because it does not have any in-flight energy sharing processes.

\section{Conclusion}
% We propose a novel Swarm-based Drone-as-a-Service framework for delivery with in-flight energy sharing. We design a novel SDaaS services composition model considering the environment's intrinsic and extrinsic constraints. We formulate the composition problem as a single objective to optimize the delivery time of consumers request. We introduce the use of support drones that share their extra available energy with the delivery drones. We propose a redundancy-based support drones number selection method to reduce the failure probability. Two approaches are presented to initially position the support drones in a flight formation. Two energy sharing methods are proposed namely, Full and Round Robin. A formation re-ordering method is presented to facilitate the energy sharing process. A modified A* algorithm is implemented to the compose the best and most optimal services from the source to the destination. Experimental results show the efficiency of our proposed approaches compared to Dijkstra's and baseline approaches. It also shows that the Full energy sharing method while positioning the support drones in the worst positions performed best in terms of successful requests compared to the other settings. It also shows that positioning the support drones in the best positions results in better delivery times. In the future work, we will expand this problem to a multi-objective problem considering the cost and the overall energy utilization.

We propose a novel Swarm-based Drone-as-a-Service framework for delivery with in-flight energy sharing. %We first design a novel swarm-based drone services composition model considering the environment's intrinsic and extrinsic constraints. 
We formulate the composition problem as a single objective to optimize a consumer's request delivery time. We then introduce the use of support drones that share their extra available energy with the delivery drones. We propose a redundancy-based method to estimate the number of support drones and reduce the failure probability. Two settings are presented to position the support drones in a flight formation initially. Two novel energy sharing methods are proposed which are based on FCFS and Round Robin approaches. A formation re-ordering method is presented to facilitate the energy sharing process. A modified A* algorithm is implemented to compose the best and optimal services from the source to the destination. Experimental results show the efficiency of our proposed approaches compared to Dijkstra's, Floyd-Warshall's, and baseline approaches. We will expand this problem to a multi-objective problem in future work to include the cost and the overall energy utilization. \textcolor{black}{We will also consider uncertainties and changes in the environment and their effects on the in-flight recharging.}
%\vspace{-8pt}

\ifCLASSOPTIONcompsoc
  \section*{Acknowledgments}
\else
  \section*{Acknowledgment}
\fi

This research was partly made possible by LE220100078 and LE180100158 grants from the Australian Research Council. The statements made herein are solely the responsibility of the authors.
%\vspace{-8pt}
%This research was partly made possible by LE220100078 grant from the Australian Research Council. The statements made herein are solely the responsibility of the authors.

\bibliographystyle{IEEEtran}
\bibliography{scholar}
\vskip -20pt plus -1fil
\begin{IEEEbiography}[{\includegraphics[width=1in,height=1.25in,clip,keepaspectratio]{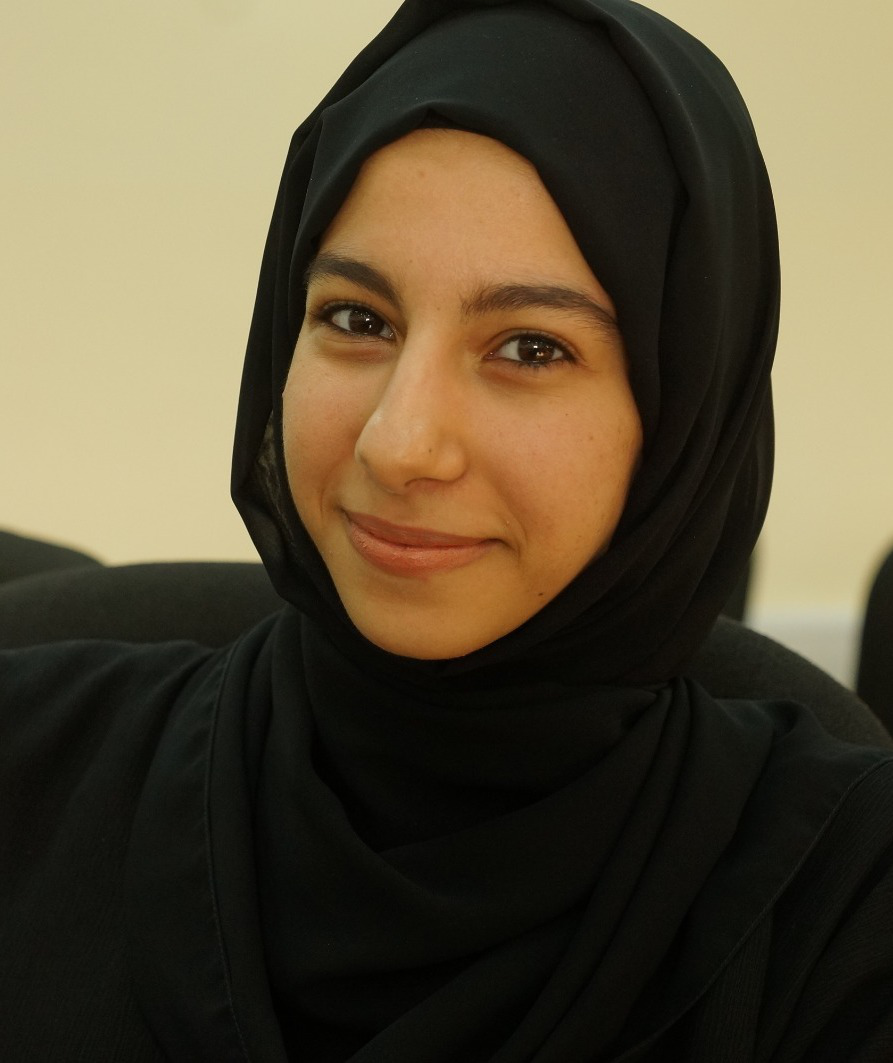}}]%
{Balsam Alkouz}
is a PhD student in the School of Computer Science at the University of Sydney. She received her bachelor's degree in IT Multimedia and her master’s degree in Computer Science from the University of Sharjah, United Arab Emirates, in 2016 and 2018 respectively. She worked as a Research Assistant in the Data Mining and Multimedia Research Group at the University of Sharjah. Her research focuses on IoT, Service Computing, and Data Mining.

\end{IEEEbiography}
\vskip -20pt plus -1fil

% \begin{IEEEbiography}[{\includegraphics[width=1in,height=1.25in,clip,keepaspectratio]{images/Amani.jpeg}}]%
% {Amani Abusafia} is a Ph.D. student under the supervision of Prof. Athman Bouguettaya at the University of Sydney.
% She received her bachelor's degree  and her master’s degree in Computer Science from the University of Sharjah, United Arab Emirates, in 2009 and 2013 respectively.
% She worked as a lecturer at the Department of Computer Science at University of Sharjah for 6 years. Her research interests include service computing, crowdsourcing, and IoT.
% \end{IEEEbiography}
\begin{IEEEbiography}[{\includegraphics[width=1in,height=1.25in,clip,keepaspectratio]{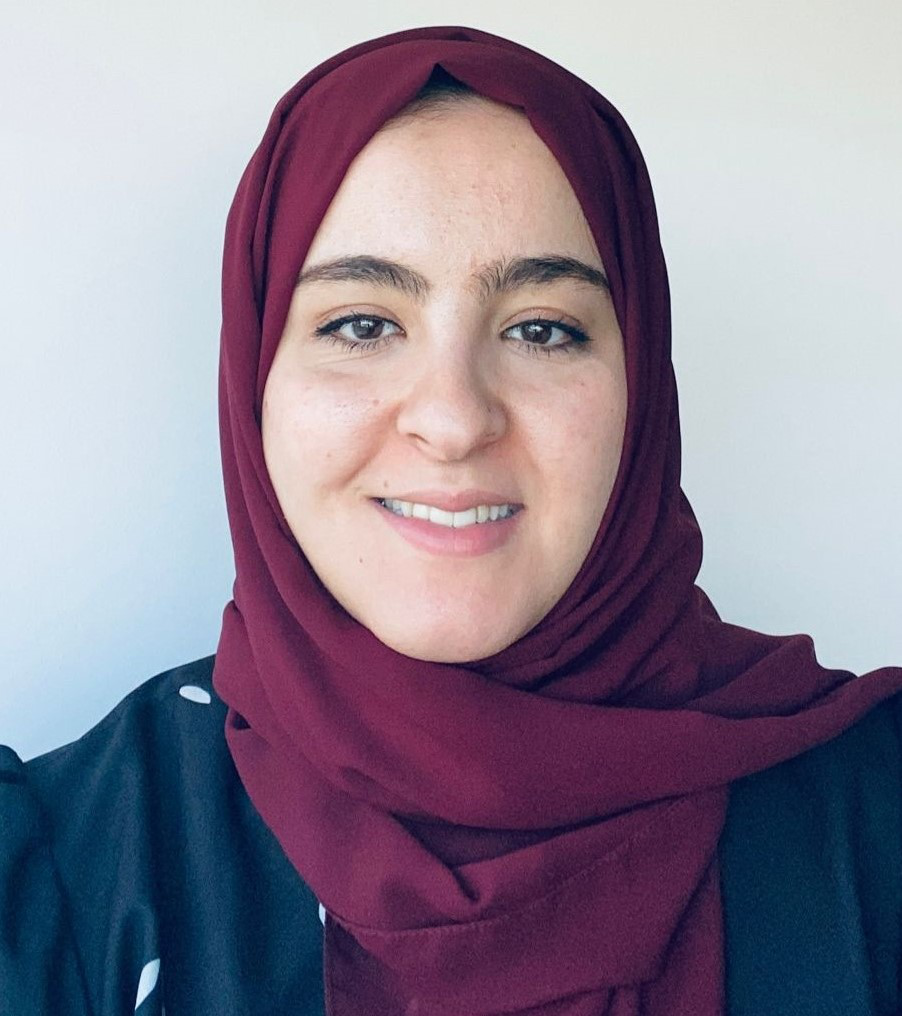}}]%
{Amani Abusafia} is a Ph.D. student in the School of Computer Science at the University of Sydney. She received her Bachelor's degree (2009) and Master's degree (2013) in Computer Science from The University of Sharjah, United Arab Emirates. She worked as a lecturer at the Department of Computer Science at the University of Sharjah for 6 years. Her research interests include Service Computing, Crowdsourcing, and IoT.\looseness=-1

\end{IEEEbiography}

\vskip -20pt plus -1fil
\begin{IEEEbiography}[{\includegraphics[width=1in,height=1.25in,clip,keepaspectratio]{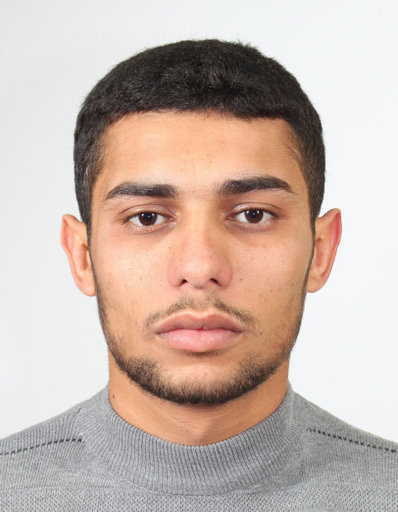}}]%
{Abdallah Lakhdari}
is a Postdoctoral Fellow in the School of Computer Science at the University of Sydney. He received his Bachelor's degree (2010) and Master's degree (2013) in Computer Science from The University of Laghouat, Algeria. He was a visiting scholar a New Mexico Tech. He worked as a lecturer at the Department of Computer Science at The University of Laghouat, Algeria. His research interests include Social Computing, Crowdsourcing, IoT, and Service Computing.
\end{IEEEbiography}
\vskip -20pt plus -1fil
% \begin{IEEEbiography}[{\includegraphics[width=1in,height=1.25in,clip,keepaspectratio]{images/athman_bouguettaya.jpeg}}]%
% {Athman Bouguettaya}
% is Professor at University of Sydney, Australia. He received his PhD in Computer Science from the University of Colorado at Boulder (USA) in 1992. He is or has been on the editorial boards of several journals including, the IEEE Transactions on Services Computing, ACM Transactions on Internet Technology, the International Journal on Next Generation Computing and VLDB Journal. He is a Fellow of the IEEE and a Distinguished Scientist of the ACM.
% \end{IEEEbiography}

\begin{IEEEbiography}[{\includegraphics[width=1in,height=1.25in,clip,keepaspectratio]{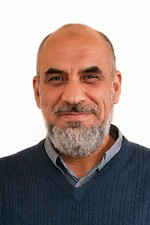}}]%
{Athman Bouguettaya}
is a Professor in the School of Computer Science at the University of Sydney. He received his Ph.D. in Computer Science from the University of Colorado at Boulder (USA) in 1992. He is or has been on the editorial boards of several journals, including the IEEE Transactions on Services Computing, ACM Transactions on Internet Technology, the International Journal on Next Generation Computing, and VLDB Journal. He is a Fellow of the IEEE and a Distinguished Scientist of the ACM. He is a member of the Academia Europaea (MAE).\looseness=-1
\end{IEEEbiography}

\end{document}